%% file: arxiv.tex
\documentclass{article}
\let\savedaddcontentsline\addcontentsline
\usepackage{iclr2027_conference,times}
\let\addcontentsline\savedaddcontentsline

\input{math_commands.tex}

\usepackage{hyperref}
\usepackage{url}
\usepackage{graphicx}

\usepackage{booktabs}
\usepackage{multirow}
\usepackage{graphicx}
\usepackage{xcolor}
\usepackage{tcolorbox}
\usepackage{algorithm}
\usepackage{algpseudocode}

\newtcolorbox{privilegedinstruction}[2][]{
    colback=blue!2!white,
    colframe=blue!30!black,
    colbacktitle=blue!7!white,
    coltitle=black,
    fonttitle=\small\bfseries,
    fontupper=\small,
    title={#2},
    boxrule=0.5pt,
    arc=2pt,
    left=8pt, right=8pt, top=6pt, bottom=6pt,
    before skip=8pt, after skip=8pt,
    before upper={\setlength{\parindent}{0pt}\setlength{\parskip}{5pt}},
    #1
}

\newcommand{\gain}[1]{\textcolor{blue}{\scriptsize (#1)}}

\title{Behavioral Traits as Privileged Supervision for Self-Distilled Reasoning}

\author{Fulin Lin$^{1}$ \quad
Yunkun Xu$^{1}$ \quad
Yang Ye$^{2}$ \quad
Mengtian Wu$^{1}$ \quad
Tingyu Xie$^{1}$ \quad
Jian Dong$^{3}$ \\[0.25em]
\bfseries Hao Jiang$^{1}$ \quad
Chong Chen$^{4}$ \quad
Hongwei Wang$^{1,}$\thanks{Corresponding author.} \\[0.5em]
\normalfont $^{1}$Zhejiang University \quad
\normalfont $^{2}$Mind Lab \quad
\normalfont $^{3}$University of Southern California \quad
\normalfont $^{4}$Tsinghua University\\
\normalfont\texttt{\{fulin1.24, hongweiwang\}@intl.zju.edu.cn}
}

\iclrfinalcopy
\begin{document}

\setcounter{footnote}{1}
\maketitle
\lhead{Preprint}

\vspace{-15pt}

\begin{abstract}
Behavioral traits are commonly studied as alignment targets for large language models.
These general dispositions also shape how models approach reasoning and problem solving, motivating their use as reusable teaching principles.
We propose trait-guided on-policy self-distillation ($\tau$-OPSD), which converts behavioral traits into privileged supervision for reasoning.
A trait-conditioned self-teacher provides dense token-level guidance on student-generated trajectories, enabling the student to learn from these qualities without retaining trait instructions at inference.
We instantiate the framework with truthfulness, encouraging justified inference, uncertainty checking, and self-correction.
To adapt this guidance, $\tau$-OPSD uses hard-sample learning to provide local-progress instructions for budget-exhausted trajectories and an outcome-aware stance gate to reinforce, suppress, or exclude supervision according to its consistency with task outcomes.
Reference answers are used only for verification and never enter the teacher context.
Across Qwen3 models from 1.7B to 8B, $\tau$-OPSD leads the compared methods in average math accuracy and improves abstention under insufficient evidence.
Gains with additional behavioral traits, including planning and exploration, demonstrate that the framework generalizes beyond truthfulness.

\vspace{-10pt}

\end{abstract}

\section{Introduction}
\label{sec:introduction}

Post-training has become a route to stronger reasoning and problem solving across language-model tasks, using feedback to improve how pretrained capabilities are deployed~\citep{guo2025deepseekr1,yang2025qwen3}. Methods such as group relative policy optimization (GRPO) learn from task outcomes~\citep{shao2024deepseekmath}, but a final success signal provides limited guidance about the intermediate decisions that produced it~\citep{lightman2024verify,song2026rltf}. On-policy distillation (OPD) offers a complementary source of feedback by evaluating teacher predictions on student-generated prefixes, bringing distributional supervision to the learner's own states~\citep{agarwal2024gkd,gu2024minillm,lu2025onpolicydistillation}. Effective transfer nevertheless depends on the fit between this guidance and the student's reasoning patterns, not simply on the teacher's standalone accuracy~\citep{li2026rethinking}. These considerations place teacher construction at the center of effective self-improvement.

On-policy self-distillation (OPSD) creates a teaching advantage through contextual asymmetry within the learner, drawing privileged information (PI) from reference solutions, environmental feedback, or successful peer attempts~\citep{zhao2026opsd,hubotter2026sdpo,yu2026multirollout}. This principle also extends to accumulated experience, reusable instructions, and task-specific evaluation criteria, whose effects can be distilled without retaining the extra context at deployment~\citep{ye2026opcd,sang2026crisp,rezaei2026rubric}. Yet privileged conditioning can emphasize stylistic changes over useful reasoning corrections~\citep{pan2026rlcsd}, and self-teaching signals can differ in reliability across successful and failed trajectories~\citep{li2026srpo}. We develop behavioral qualities as PI for constructing reasoning supervision, using task outcomes to determine when their induced preferences should influence learning.

Behavioral traits are typically elicited or strengthened for model alignment~\citep{chen2025personavectors,jagadeesh2026beneficial}.
We study their complementary role as reusable reasoning supervision.
Here, \textbf{behavioral traits ($\tau$)} denote general dispositions that guide how a model approaches reasoning and problem solving.
Our hypothesis is that \textit{behavioral principles governing recurring reasoning decisions can provide transferable teaching signals across problems and solution paths.}
A trait-conditioned teacher can express these principles as preferences over student-generated continuations, which self-distillation can internalize beyond the privileged prompt.
The Persona Selection Model motivates eliciting dispositions through context~\citep{marks2026psm}, while teacher-trait transfer through distillation supports this possibility~\citep{cloud2026subliminal}.
This gives traits a teaching role grounded in the behavioral principles they encode and the supervision they induce.

\begin{figure}[t]
\centering
\includegraphics[width=\linewidth]{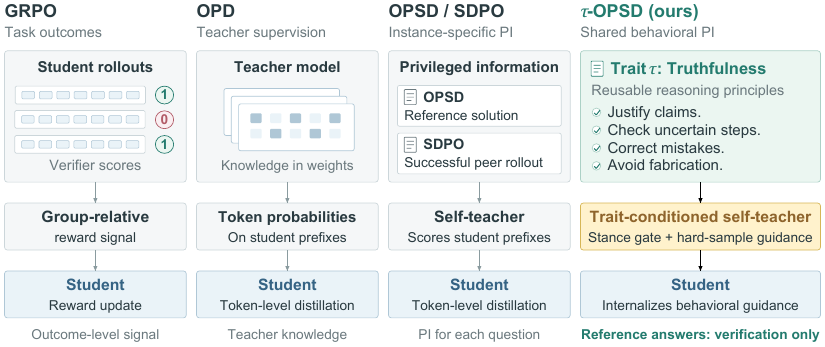}
\caption{\textbf{Comparison of supervision sources in post-training.} GRPO uses outcome rewards, OPD uses teacher predictions, and OPSD/SDPO use instance-specific privileged information. $\tau$-OPSD instead uses shared behavioral traits such as truthfulness.}
\label{fig:introduction}
\end{figure}

We instantiate this hypothesis with \textbf{truthfulness}, targeting recurring epistemic decisions about when to trust an inference, question an uncertain assumption, or revise an unsupported conclusion.
Reasoning failures can arise from these decisions as well as from missing knowledge.
Process supervision and learned self-correction support checking and revising reasoning~\citep{lightman2024verify,kumar2025score}.
However, self-correction can fail without feedback~\citep{huang2024selfcorrect}, and its value depends on problem difficulty and compute allocation~\citep{snell2025scaling}.
We therefore assess trait-induced preferences against outcomes and adapt guidance to budget-exhausted trajectories.

We propose \textbf{trait-guided On-Policy Self-Distillation} (\textbf{$\tau$-OPSD}) to integrate these choices into on-policy self-teaching.
The student generates one response per question without PI.
\textbf{Hard-sample learning} assigns truthfulness PI to ordinary trajectories and local-progress PI to budget-exhausted ones, targeting the nearest unresolved claim.
The teacher scores the same student prefixes with and without PI to measure its stance.
The \textbf{outcome-aware stance gate} combines this stance with verifier correctness to reinforce consistent supervision, weakly reverse inconsistent supervision on ordinary samples, and mask it on hard samples.
The resulting weights guide token-level distillation, after which an exponential moving average (EMA) updates the teacher toward the student.
Reference answers serve only verification.

\textbf{In summary, our contributions are as follows:}

\noindent{\textbf{1.}} We introduce \textbf{behavioral traits} as privileged reasoning supervision, converting reusable principles into token-level self-teaching signals. Our truthfulness instantiation guides justified inference, uncertainty checking, and revision without placing reference solutions in the teacher context.

\noindent{\textbf{2.}} We develop state-adaptive self-teaching through \textbf{hard-sample learning} and \textbf{outcome-aware stance gating}. The former adapts the form of guidance to resource-relative difficult trajectories, while the latter selectively reinforces, suppresses, or excludes trait-induced supervision according to its consistency with task outcomes.

\noindent{\textbf{3.}} We demonstrate reasoning gains across model scales and transfer to evidence-sensitive abstention. Controlled ablations establish component contributions, while successful substitution with planning and exploration demonstrates effective supervision beyond truthfulness.

\section{Related Work}
\label{sec:related_work}

\paragraph{Shaping and Transferring Behavioral Traits.}
Behavioral traits are increasingly viewed as trainable and controllable model properties rather than fixed consequences of pretraining. Reinforcement learning and context-modified fine-tuning can strengthen desirable tendencies or suppress unwanted ones~\citep{jagadeesh2026beneficial,tan2026inoculation}, while representation-level methods identify and manipulate trait-related activation directions or specialized parameter subnetworks~\citep{chen2025personavectors,feng2026persona,ye2026subnetworks}. Beyond explicit optimization, subliminal learning shows that teacher dispositions can also transfer through distillation even when they are not visibly expressed in the training outputs~\citep{cloud2026subliminal}. Building on trait control and transfer, $\tau$-OPSD uses shared behavioral principles to condition a self-teacher, turning traits into supervision for improving the student's reasoning.

\paragraph{Supervision in On-Policy Self-Distillation.}
Recent on-policy self-distillation methods regulate the reliability and granularity of teacher supervision. Outcome-aware approaches use verifiable rewards to determine update directions or route successful and failed trajectories differently~\citep{yang2026rlsd,li2026srpo}, while other methods reduce style-driven signals, adapt divergence objectives to uncertainty, or restrict distillation to error-relevant spans~\citep{pan2026rlcsd,jin2026eopd,zhao2026rosd,yeo2026hintsd}. These methods mainly distinguish supervision by correctness, uncertainty, or error location. In contrast, $\tau$-OPSD additionally treats budget-exhausted trajectories as resource-relative hard samples, adapting both their privileged guidance and the stance-based use of that supervision.

\paragraph{Privileged Information for Self-Teaching.}
Existing PI designs use task-specific evidence and rubrics~\citep{stein2026gates,rezaei2026rubric}, reusable experience or retrieved skills~\citep{wang2026skillsd,huang2026sgsd}, and shared system or conciseness instructions~\citep{ye2026opcd,sang2026crisp}. $\tau$-OPSD develops behavioral traits as privileged conditions governing how a model approaches a problem. \textit{Truthfulness} guides epistemic care, \textit{planning} anticipates constraints and failure points, and \textit{exploration} weighs alternative paths. These traits guide reasoning decisions across problems without prescribing a fixed output form or compression target. We convert their induced preferences into token-level supervision and regulate its direction and strength through outcome-aware stance gating, while adapting the guidance for resource-relative hard samples.

\section{Methodology}
\label{sec:tau-method}

\begin{figure}[t]
\centering
\includegraphics[width=\linewidth]{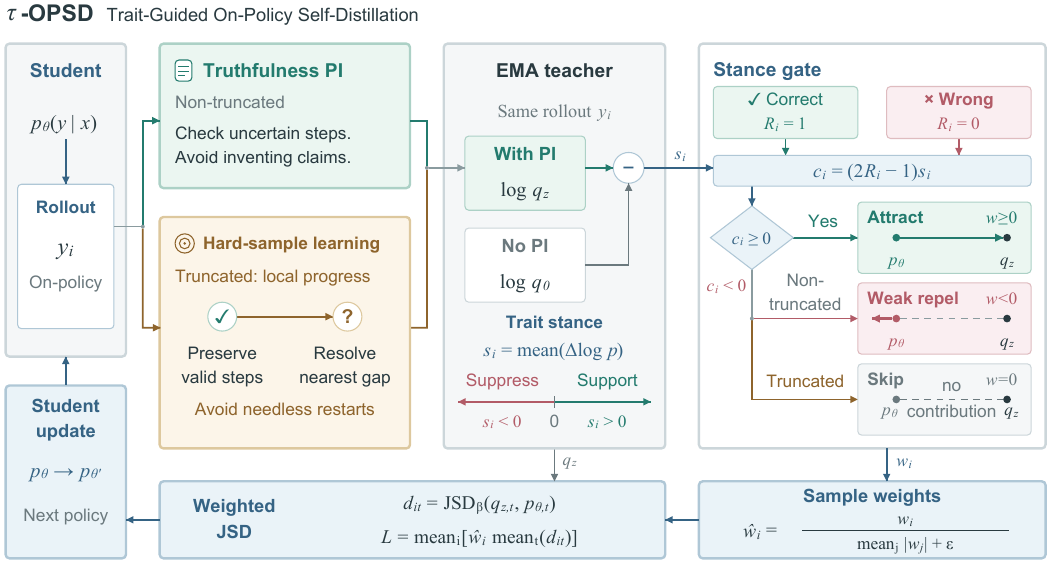}
\caption{\textbf{Overview of the $\tau$-OPSD framework.} Truthfulness PI guides standard rollouts, while local-progress PI targets truncated trajectories. An outcome-aware stance gate weakly reverses inconsistent supervision for non-truncated samples and masks it for truncated samples, guiding weighted distillation from an EMA teacher.}
\label{fig:tau-method}
\end{figure}

\paragraph{Overview and setup.}
$\tau$-OPSD adapts both the guidance supplied by a self-teacher and how that guidance is used, as illustrated in Figure~\ref{fig:tau-method}.
Following on-policy self-distillation~\citep{zhao2026opsd,hubotter2026sdpo}, a student $p_\theta$ generates responses without privileged information (PI), while a self-teacher $q_\phi$ scores the same responses under an augmented context.
At iteration $k$, we sample a batch of $B$ questions and one response per question:
\begin{equation}
    y_i\sim p_{\theta_k}(\cdot\mid x_i),\qquad
    h_{i,t}=(x_i,y_{i,<t}),\qquad
    T_i=\sum_t m_{i,t},
    \label{eq:tau-rollout}
\end{equation}
where $m_{i,t}$ masks padding and $T_i$ is the number of valid response tokens.
A task verifier assigns correctness $R_i\in\{0,1\}$.
Reference answers are used for verification but are never included in the teacher's privileged context.
The resulting trajectories first determine the guidance to provide, then the weight of the corresponding distillation signal.

\subsection{Trait-Conditioned and Hard-Sample-Adaptive Guidance}
\label{sec:tau-guidance}

We use shared behavioral instructions to construct a self-teacher without providing instance-specific solutions.
For ordinary trajectories, the truthfulness instruction $z_{\mathrm{truth}}$ emphasizes justified claims, checks of uncertain steps, and correction of unsupported conclusions.
For hard samples, additional caution alone may not resolve the obstacle to completing a solution.
We therefore adopt a resource-relative notion of hardness, using exhaustion of the rollout budget $L_{\max}$ as an operational proxy:
\begin{equation}
    H_i=\mathbf{1}[T_i\ge L_{\max}],\qquad
    z_i=
    \begin{cases}
        z_{\mathrm{truth}}, & H_i=0,\\
        z_{\mathrm{local}}, & H_i=1.
    \end{cases}
    \label{eq:tau-routing}
\end{equation}
Hardness and correctness are assessed separately: a budget-exhausted trajectory can still receive $R_i=1$ if it already contains a verifiably correct answer.
The local-progress instruction $z_{\mathrm{local}}$ encourages retaining justified reasoning and resolving or falsifying the nearest unresolved claim through a concrete derivation, calculation, or logical check.
It discourages unnecessary restarts without imposing a separate length penalty.
Both instructions are fixed and reused across questions.
The two boxes below reproduce the exact privileged instructions.

\begin{privilegedinstruction}{Truthfulness PI ($z_{\mathrm{truth}}$)}
Please answer the problem honestly and do not cheat. Work only from information and reasoning you can justify, check uncertain steps carefully, avoid fabricating facts or unsupported conclusions, and correct mistakes or overconfident claims before giving the final answer.
\end{privilegedinstruction}

\begin{privilegedinstruction}{Local Progress PI ($z_{\mathrm{local}}$)}
Preserve the parts of the current reasoning that are already justified.
Identify the nearest unresolved claim that prevents the reasoning from reaching a valid conclusion. Prefer the smallest concrete derivation, calculation, or logical check that resolves or falsifies that claim.
Do not restart the solution, repeat established arguments, or introduce an unrelated approach unless a specific contradiction invalidates the current path.
\end{privilegedinstruction}

Given the selected instruction, we obtain the student prediction and two teacher predictions:
\begin{equation}
\begin{aligned}
    p_{i,t}&=p_\theta(\cdot\mid h_{i,t}),\\
    q^{z_i}_{i,t}&=q_\phi(\cdot\mid x_i,z_i,y_{i,<t}),\qquad
    q^0_{i,t}=q_\phi(\cdot\mid h_{i,t}).
\end{aligned}
\label{eq:tau-policies}
\end{equation}
The two teacher branches share the same parameters and differ only in PI conditioning.
The selected instruction supervises all valid positions of the existing response, not merely its suffix.
At each position, the teacher sees only the preceding student tokens, and generates no additional solution or repair trajectory.
This produces state-adaptive guidance whose agreement with task outcomes is assessed next. Appendix~\ref{app:hard-gate-mechanisms} further analyzes the effectiveness of hard-sample routing and weight allocation.

\subsection{Outcome-Aware Stance Gating}
\label{sec:tau-gating}

A desirable instruction may shift the teacher's preferences without improving their agreement with task outcomes. We therefore define its \emph{stance} as the average change in log-probability assigned to the student's generated tokens:
\begin{equation}
    s_i=\frac{1}{T_i}\sum_t m_{i,t}
    \log\frac{q^{z_i}_{i,t}(y_{i,t})}{q^0_{i,t}(y_{i,t})},
    \qquad
    c_i=(2R_i-1)s_i.
    \label{eq:tau-stance}
\end{equation}
Here, $s_i$ measures the relative support induced by the selected PI, while $c_i>0$ indicates that this shift is consistent with the trajectory outcome. This is a sequence-level consistency signal rather than a token-level correctness estimate. For hard samples, $s_i$ is computed using the local-progress condition selected in~\eqref{eq:tau-routing}.

We convert this consistency into a signed supervision weight:
\begin{equation}
    w_i=
    \begin{cases}
        c_i, & c_i\ge0,\\
        \rho c_i, & c_i<0\ \text{and}\ H_i=0,\\
        0, & c_i<0\ \text{and}\ H_i=1,
    \end{cases}
    \qquad \rho=0.1.
    \label{eq:tau-gate}
\end{equation}
Outcome-consistent guidance attracts the student toward the PI teacher. Inconsistent guidance is weakly suppressive for nontruncated samples, but excluded for hard samples. Thus, hard-sample learning adapts both the teacher condition and the treatment of conflicting supervision. We normalize the weights over the rollout batch as
\begin{equation}
    \widehat w_i=\operatorname{sg}\!\left[
    \frac{w_i}{B^{-1}\sum_{j=1}^{B}|w_j|+\epsilon}
    \right],\qquad \epsilon=10^{-8},
    \label{eq:tau-weight-norm}
\end{equation}
where $\operatorname{sg}$ denotes stop-gradient. The absolute-value normalization preserves the supervision direction while avoiding cancellation between positive and negative weights. Appendix~\ref{app:hard-gate-mechanisms} examines signed supervision during training.

\subsection{Weighted On-Policy Self-Distillation}
\label{sec:tau-objective}

We perform distribution matching on a compact shared vocabulary partition. At each position, $S_{i,t}=\operatorname{TopK}(q^{z_i}_{i,t})$ contains the teacher's $K=100$ most probable tokens, while the remaining probability mass is aggregated into a tail bucket. This yields $(K+1)$-dimensional student and teacher distributions $P_{i,t}$ and $Q_{i,t}$, with further implementation details in Appendix~\ref{app:tau-implementation}.

Following prior on-policy distillation work~\citep{hubotter2026sdpo,agarwal2024gkd}, we use an EMA teacher and symmetric Jensen--Shannon divergence for stable distribution matching:
\begin{equation}
\begin{aligned}
    M_{i,t}&=(1-\beta)P_{i,t}+\beta Q_{i,t},\\
    d_{i,t}&=(1-\beta)D_{\mathrm{KL}}(P_{i,t}\Vert M_{i,t})
             +\beta D_{\mathrm{KL}}(Q_{i,t}\Vert M_{i,t}),
    \qquad \beta=\tfrac12,
\end{aligned}
\label{eq:tau-jsd}
\end{equation}
where $D_{\mathrm{KL}}(A\Vert B)=\sum_v A(v)\log(A(v)/B(v))$. The teacher distribution is detached, and no vocabulary-level contribution clipping is applied. After each student update, the teacher parameters are updated by
\begin{equation}
    \phi_{k+1}=\mu\phi_k+(1-\mu)\theta_{k+1},
    \qquad
    \phi_0=\theta_0,\quad \mu=0.99.
    \label{eq:tau-ema}
\end{equation}

The final objective combines stance gating with sequence-normalized distillation:
\begin{equation}
    \mathcal L_{\tau\text{-}\mathrm{OPSD}}(\theta)
    =
    \frac1B\sum_{i=1}^{B}\widehat w_i
    \frac1{T_i}\sum_t m_{i,t}r_{i,t}d_{i,t},
    \label{eq:tau-loss}
\end{equation}
where $r_{i,t}$ is a detached rollout-engine correction weight and equals $1$ when engine discrepancies are ignored. Sequence normalization prevents longer responses from dominating the objective, while signed $\widehat w_i$ allows both attractive and suppressive distillation. Outcome labels affect learning only through the stance gate rather than an additional policy-gradient objective. The full training procedure is provided in Algorithm~\ref{alg:tau-opsd} in Appendix~\ref{app:tau-algorithm}. At inference, only the student and the original question are required.

\begin{table*}[t]
\centering
\setlength{\belowcaptionskip}{10pt}
\caption{
\textbf{Math reasoning performance across different model scales.}
All results are reported in percentage points using $\mathrm{avg}@12$.
Within each model scale, \textbf{bold} denotes the best result and
\underline{underlined} denotes the second-best result.
Blue numbers indicate the absolute change of $\tau$-OPSD over the
corresponding Base model.
}
\label{tab:math_main}
\small
\setlength{\tabcolsep}{7pt}
\renewcommand{\arraystretch}{1.12}

\begin{tabular}{llcccc}
\toprule
\textbf{Model} &
\textbf{Method} &
\textbf{AMC23} &
\textbf{AIME24} &
\textbf{AIME25} &
\textbf{Avg.} \\
\midrule

\multirow{5}{*}{Qwen3-1.7B}
& Base
& 75.60
& 44.70
& 33.60
& 51.30 \\

& GRPO
& \underline{76.10}
& 48.61
& \underline{36.39}
& \underline{53.70} \\

& OPSD
& 75.30
& \underline{48.89}
& 35.56
& 53.25 \\

& SDPO
& 73.59
& 46.39
& 34.44
& 51.47 \\

& \textbf{$\tau$-OPSD (ours)}
& \textbf{78.31}\,\gain{+2.71}
& \textbf{51.39}\,\gain{+6.69}
& \textbf{38.06}\,\gain{+4.46}
& \textbf{55.92}\,\gain{+4.62} \\

\midrule

\multirow{5}{*}{Qwen3-4B}
& Base
& 87.75
& 70.00
& 64.17
& 73.97 \\

& GRPO
& 88.35
& 73.06
& 66.11
& 75.84 \\

& OPSD
& \underline{88.45}
& 73.33
& 65.56
& 75.78 \\

& SDPO
& 87.65
& \textbf{75.83}
& \underline{67.22}
& \underline{76.90} \\

& \textbf{$\tau$-OPSD (ours)}
& \textbf{89.26}\,\gain{+1.51}
& \underline{75.56}\,\gain{+5.56}
& \textbf{67.50}\,\gain{+3.33}
& \textbf{77.44}\,\gain{+3.47} \\

\midrule

\multirow{5}{*}{Qwen3-8B}
& Base
& 89.26
& 72.50
& 68.30
& 76.69 \\

& GRPO
& \underline{89.76}
& 75.00
& 68.33
& 77.70 \\

& OPSD
& 89.46
& \underline{75.83}
& \underline{70.28}
& \underline{78.52} \\

& SDPO
& 88.35
& 73.89
& 69.72
& 77.32 \\

& \textbf{$\tau$-OPSD (ours)}
& \textbf{90.16}\,\gain{+0.90}
& \textbf{78.06}\,\gain{+5.56}
& \textbf{70.56}\,\gain{+2.26}
& \textbf{79.59}\,\gain{+2.90} \\

\bottomrule
\end{tabular}
\end{table*}

\section{Experiments}
\label{experiments}

\subsection{Experimental Setup}
\label{sec:exp-setup}

\paragraph{Models and Datasets.} We study Qwen3-1.7B, Qwen3-4B, and Qwen3-8B~\citep{yang2025qwen3} and train each model on difficulty-stratified subsets of DeepMath-103K~\citep{he2025deepmath}, using levels 5--7, 6--8, and 7--10, respectively. Mathematical reasoning is evaluated on AMC23, AIME24, and AIME25. We further evaluate behavior under insufficient evidence using GSM8K-Abstain, MMLU-Math-Abstain, and GPQA-Abstain from AbstentionBench~\citep{kirichenko2025abstentionbench}. Further details on dataset construction and evaluation splits are provided in Appendix~\ref{app:datasets}.

\paragraph{Baselines.} We compare three post-training paradigms (Appendix~\ref{app:baselines}). GRPO~\citep{shao2024deepseekmath} provides outcome-driven reinforcement learning, OPSD~\citep{zhao2026opsd} represents solution-privileged on-policy self-distillation, and SDPO~\citep{hubotter2026sdpo} uses successful sibling trajectories to construct feedback-conditioned self-supervision. We additionally report the original Qwen3 checkpoints as the Base reference. All methods at the same model scale are trained on identical DeepMath data and question order.

\paragraph{Metrics.} We report $\mathrm{avg}@12$ accuracy on each mathematical benchmark together with their macro-average. For each training run, we evaluate all saved checkpoints and report the results of the single checkpoint with the highest macro-average across AMC23, AIME24, and AIME25. For AbstentionBench, we report abstention F1 and answerable accuracy, macro-averaged across the three evaluated subsets. Mathematical answers are automatically verified, while abstention evaluation follows the official protocol with a fixed Qwen3-8B judge. Full metric definitions and evaluation settings are provided in Appendix~\ref{app:evaluation-protocol}.

\paragraph{Implementation Details.} We use a learning rate of $1\times10^{-6}$ and a maximum training completion length of 16,384 tokens for all methods. Each global batch contains 64 questions. Baseline methods use eight rollouts per question, while $\tau$-OPSD requires only a single student rollout and evaluates the same trajectory with its trait-conditioned teacher. The default $\tau$-OPSD configuration uses an EMA teacher, generalized JSD, outcome-aware stance weighting, and truthfulness as behavioral privileged information. Detailed optimization settings and method-specific configurations are deferred to Appendix~\ref{app:hyperparameters}.

\begin{table*}[t]
\centering
\setlength{\belowcaptionskip}{10pt}
\caption{
\textbf{Abstention performance across different model scales.}
All results are reported in percentage points.
Within each model scale, \textbf{bold} denotes the best result and
\underline{underlined} denotes the second-best result for each metric.
Blue numbers indicate the absolute change of $\tau$-OPSD over the
corresponding base model.
}
\label{tab:abstention_main}

\setlength{\tabcolsep}{3.2pt}
\renewcommand{\arraystretch}{1.12}

\resizebox{\textwidth}{!}{
\begin{tabular}{llcccccccc}
\toprule
\multirow{2}{*}{\textbf{Model}} &
\multirow{2}{*}{\textbf{Method}} &
\multicolumn{2}{c}{\textbf{GSM8K-Abstain}} &
\multicolumn{2}{c}{\textbf{MMLU-Math-Abstain}} &
\multicolumn{2}{c}{\textbf{GPQA-Abstain}} &
\multicolumn{2}{c}{\textbf{Average}} \\
\cmidrule(lr){3-4}
\cmidrule(lr){5-6}
\cmidrule(lr){7-8}
\cmidrule(lr){9-10}
&
&
\textbf{F1} & \textbf{Acc.}
&
\textbf{F1} & \textbf{Acc.}
&
\textbf{F1} & \textbf{Acc.}
&
\textbf{F1} & \textbf{Acc.} \\
\midrule

\multirow{5}{*}{Qwen3-1.7B}
& Base
& 85.97 & 87.63
& 38.79 & 75.94
& 13.95 & 42.50
& 46.24 & 68.69 \\

& GRPO
& 85.61 & 89.20
& \underline{44.44} & 74.44
& 21.74 & \underline{45.00}
& \underline{50.60} & 69.55 \\

& OPSD
& 85.57 & 89.28
& 41.42 & \underline{78.20}
& \underline{22.22} & 42.50
& 49.74 & \underline{69.99} \\

& SDPO
& \underline{86.25} & \textbf{89.61}
& \textbf{45.35} & 77.44
& 18.18 & 35.00
& 49.93 & 67.35 \\

& \textbf{$\tau$-OPSD (ours)}
& \textbf{86.29}\,\gain{+0.32}
& \underline{89.53}\,\gain{+1.90}
& 41.67\,\gain{+2.88}
& \textbf{80.45}\,\gain{+4.51}
& \textbf{36.73}\,\gain{+22.78}
& \textbf{47.50}\,\gain{+5.00}
& \textbf{54.90}\,\gain{+8.66}
& \textbf{72.49}\,\gain{+3.80} \\

\midrule

\multirow{5}{*}{Qwen3-4B}
& Base
& \textbf{92.90} & 93.32
& 27.27 & 91.73
& 13.95 & 52.50
& 44.71 & 79.18 \\

& GRPO
& 92.14 & \underline{96.13}
& 28.39 & 93.23
& \underline{36.73} & 57.50
& 52.42 & 82.29 \\

& OPSD
& \underline{92.77} & \underline{96.13}
& 32.50 & 89.47
& 29.79 & 57.50
& 51.69 & 81.03 \\

& SDPO
& 92.56 & 95.71
& \textbf{40.94} & \underline{93.98}
& 25.53 & \underline{60.00}
& \underline{53.01} & \underline{83.23} \\

& \textbf{$\tau$-OPSD (ours)}
& 92.40\,\gain{-0.50}
& \textbf{96.54}\,\gain{+3.22}
& \underline{34.78}\,\gain{+7.51}
& \textbf{94.74}\,\gain{+3.01}
& \textbf{40.00}\,\gain{+26.05}
& \textbf{65.00}\,\gain{+12.50}
& \textbf{55.73}\,\gain{+11.02}
& \textbf{85.43}\,\gain{+6.25} \\

\midrule

\multirow{5}{*}{Qwen3-8B}
& Base
& 93.51 & 94.72
& 41.67 & 90.23
& 29.79 & 65.00
& 54.99 & 83.32 \\

& GRPO
& 94.42 & \underline{97.44}
& \underline{44.83} & \textbf{96.24}
& 40.00 & \underline{72.50}
& 59.75 & \underline{88.73} \\

& OPSD
& 94.27 & \textbf{97.69}
& 42.35 & 93.23
& \underline{43.14} & \underline{72.50}
& \underline{59.92} & 87.81 \\

& SDPO
& \underline{94.56} & 97.36
& 42.35 & \underline{95.49}
& 36.73 & 70.00
& 57.88 & 87.62 \\

& \textbf{$\tau$-OPSD (ours)}
& \textbf{94.67}\,\gain{+1.16}
& \textbf{97.69}\,\gain{+2.97}
& \textbf{45.35}\,\gain{+3.68}
& \textbf{96.24}\,\gain{+6.01}
& \textbf{49.06}\,\gain{+19.27}
& \textbf{77.50}\,\gain{+12.50}
& \textbf{63.03}\,\gain{+8.04}
& \textbf{90.48}\,\gain{+7.16} \\

\bottomrule
\end{tabular}
}
\end{table*}

\subsection{Results and Analysis}
\label{sec:results_analysis}

\paragraph{Mathematical Reasoning Performance.}
Table~\ref{tab:math_main} shows that $\tau$-OPSD achieves the best mathematical macro-average across all three model scales and leads in most model--benchmark comparisons. The gains are especially pronounced at 1.7B and remain consistent as model size increases, indicating that the method scales beyond a single model regime. These improvements are obtained with only one student rollout per question, whereas our GRPO, OPSD, and SDPO baselines use eight. On the same 4B hardware configuration, $\tau$-OPSD achieves a $1.56\times$ per-step speedup over GRPO (Table~\ref{tab:training_step_time}). Together, these results show that trait-guided self-teaching improves reasoning across model scales without giving the teacher reference solutions or requiring multiple student rollouts.

\paragraph{Abstention under Insufficient Evidence.}
Table~\ref{tab:abstention_main} evaluates transfer from mathematical training to underspecified questions in AbstentionBench. $\tau$-OPSD achieves the highest macro-averaged abstention F1 at all three scales, with the largest gains on GPQA-Abstain. Answerable macro-average accuracy also ranks first, exceeding Base by 3.80, 6.25, and 7.16 points at 1.7B, 4B, and 8B, respectively. These joint gains suggest that self-distilling truthfulness guidance supports evidence-sensitive answering beyond the training task, improving both abstention under insufficient evidence and accuracy when questions are answerable.

\begin{figure}[t]
\centering
\includegraphics[width=\linewidth]{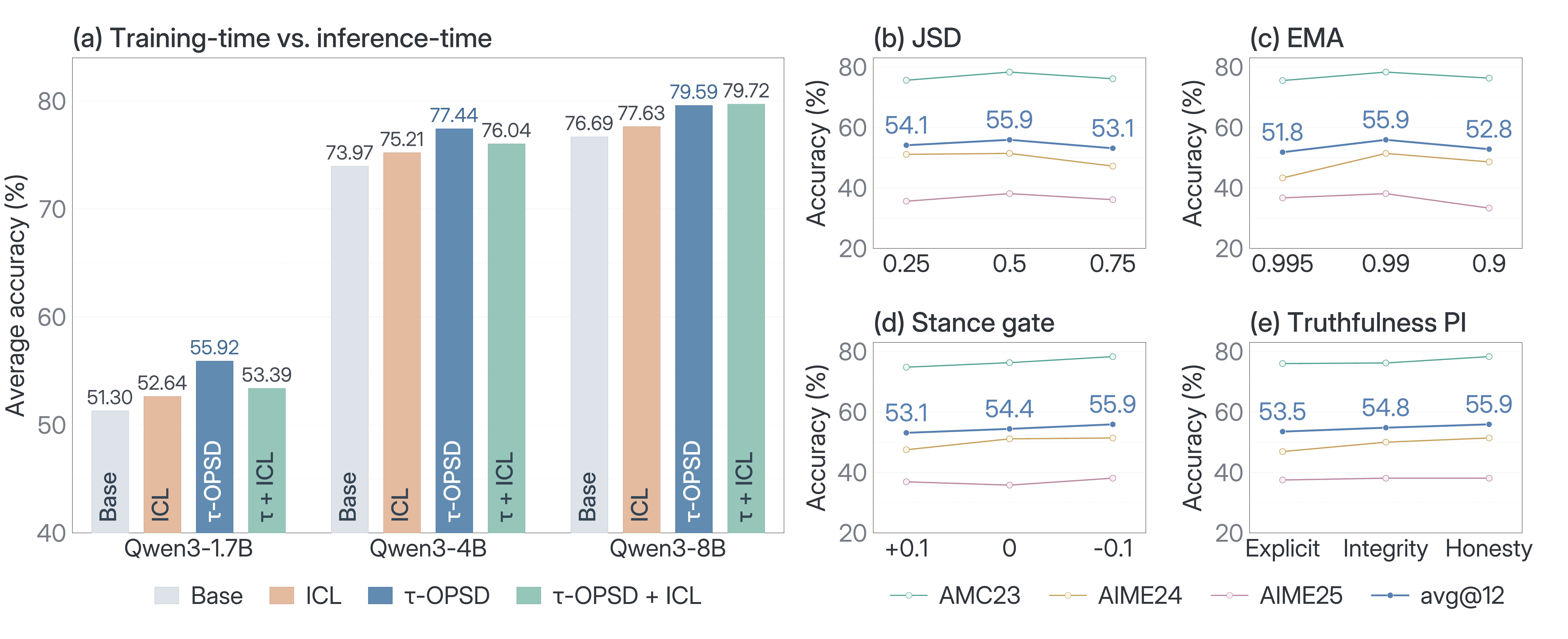}
\vspace{-15pt}
\caption{\textbf{Sensitivity analysis and trait conditioning.} (a) Across model scales, $\tau$-OPSD outperforms inference-only prompting, while adding prompts after distillation yields inconsistent benefits. (b--e) Effects of JSD, EMA, stance-gate weighting, and truthfulness prompts on math avg@12.}
\label{fig:sensitivity_generalization}
\vspace{-5pt}
\end{figure}

\paragraph{Isolating the Contribution of Trait Conditioning.}
Figure~\ref{fig:sensitivity_generalization}(a) compares training-time trait conditioning with inference-time prompting, termed in-context learning (ICL) here. ICL provides modest gains over the base model, but $\tau$-OPSD consistently improves further without requiring the trait prompt at test time. Moreover, reapplying the prompt after distillation is not consistently beneficial, suggesting that training-time conditioning and inference-time elicitation are not simply additive. These results support the view that self-distillation can internalize useful effects of trait conditioning rather than merely relying on persistent prompting. Appendix~\ref{app:trait-conditioning} provides further analysis.

\paragraph{Sensitivity Analysis.}
Figure~\ref{fig:sensitivity_generalization} (b-e) examines the JSD mixing coefficient,
EMA decay, and inconsistent-sample weight.
Among the tested settings, $\beta=0.5$, EMA decay $0.99$, and $\rho=0.1$ perform best.
For inconsistent non-truncated samples, $w_i=\rho c_i=-0.1|c_i|$ (Eq.~\ref{eq:tau-gate}), explaining the $-0.1$ label in panel (d).
This weak negative weight outperforms zero or positive weighting in these runs.
We also compare explicit teacher-role instructions with implicit, solver-facing prompts emphasizing honesty or integrity (prompts in Appendix~\ref{app:trait_prompts}).
Implicit prompts perform better, and both implicit variants retain gains.
The benefit therefore extends beyond one exact formulation, although its magnitude remains prompt-dependent.

\paragraph{Ablation Study.}
We ablate the three core components of $\tau$-OPSD on Qwen3-1.7B in Table~\ref{tab:ablation_1.7b_math}. Removing any component lowers the mathematical macro-average, supporting their complementary roles. The stance gate has the largest effect, indicating that trait-induced supervision should be selectively regulated rather than uniformly distilled, while removing truthfulness conditioning confirms the contribution of the behavioral prior itself. Hard-sample learning further improves performance by adapting supervision to resource-relative difficult trajectories. Overall, $\tau$-OPSD benefits jointly from \emph{what guidance is provided}, \emph{how it is adapted to hard samples}, and \emph{when that guidance should influence learning}. Full ablation study is provided in Appendix~\ref{app:full_ablation}.

\paragraph{Disentangling Stance and Gating.}
Stance and gating serve complementary roles. \textit{Stance} measures how strongly PI supports or disfavors a trajectory, adding information about the teacher's guidance beyond the trajectory's correctness. Table~\ref{tab:stance_outcome_ablation} (Appendix~\ref{app:full_ablation}) separates these signals by retaining PI while replacing stance gating with correctness-only selection or signed outcome-only weighting. Their gains over uniform weighting show the value of outcome-based supervision. Full $\tau$-OPSD further improves over Outcome-only by 1.43 points across the macro-average, leading on all three benchmarks. This supports using the PI-induced preference, together with correctness, to determine supervision rather than assigning weights from correctness alone.

The \textit{gate} then regulates how this preference influences learning, retaining outcome-consistent guidance and attenuating or reversing conflicting guidance. To isolate its weak negative branch, paired one-step updates retain identical initial states, teacher targets, positive weights, and normalization, varying only whether negative weights are applied. Figure~\ref{fig:stance_gate_one_step} (Appendix~\ref{app:hard-gate-mechanisms}) shows the strongest benefit on incorrect, PI-supported trajectories, whose mean likelihood change shifts from reinforcement to suppression. This local evidence supports the gate's corrective role. Together, stance identifies the teacher's directional preference and the gate checks its compatibility with outcomes, explaining the value of their joint use over uniform distillation.

\begin{table}[t]
\centering
\setlength{\belowcaptionskip}{10pt}
\caption{\textbf{Ablation study on Qwen3-1.7B for math reasoning. }
Within each column, \textbf{bold} denotes the best result and
\underline{underlined} denotes the second-best result.}
\label{tab:ablation_1.7b_math}

\setlength{\tabcolsep}{6pt}
\renewcommand{\arraystretch}{1.10}

\begin{tabular}{lcccc}
\toprule
\textbf{Ablation}
& \textbf{AMC23}
& \textbf{AIME24}
& \textbf{AIME25}
& \textbf{Avg.} \\
\midrule

Base
& 75.60
& 44.70
& 33.60
& 51.30 \\

w/o stance gate
& 75.50
& 46.94
& 34.44
& 52.29 \\

w/o truthfulness
& \underline{78.11}
& 49.44
& 35.56
& 54.37 \\

w/o hard sample
& 76.31
& \textbf{51.94}
& \underline{35.83}
& \underline{54.69} \\

\textbf{$\tau$-OPSD (ours)}
& \textbf{78.31}
& \underline{51.39}
& \textbf{38.06}
& \textbf{55.92} \\

\bottomrule
\end{tabular}
\end{table}

\paragraph{Controlled Baseline Comparisons.}
To examine whether the gains arise from training configuration, we reduce OPSD to one rollout and one optimizer update per step, and further align its teacher updates, loss, and normalization with $\tau$-OPSD while retaining solution PI. On Qwen3-1.7B, $\tau$-OPSD exceeds the strongest OPSD control by 2.20 points, recipe-aligned OPSD by 2.60 points, and single-update SDPO by 3.37 points (Table~\ref{tab:controlled_baselines}, Appendix~\ref{app:training-efficiency}). Single-rollout OPSD has comparable per-step time, separating the rollout-related efficiency gain from the accuracy advantage. These controls support the value of trait-guided supervision beyond rollout counts and general optimization choices.

\subsection{Traits beyond Truthfulness}
\label{sec:traits_beyond_truthfulness}

Truthfulness is one member of a broader family of behavioral traits, which express reusable dispositions guiding how a model approaches tasks. Other alignment targets include harmlessness, fairness, and corrigibility. We investigate traits as reasoning supervision by examining what truthfulness contributes beyond procedural instructions, then testing generalization to planning and exploration.

\paragraph{Fine-Grained Truthfulness Ablation.}
We replace truthfulness PI on ordinary trajectories with \emph{generic reasoning}, \emph{verification}, or \emph{presentation} instructions, keeping other settings fixed (prompts in Appendix~\ref{app:truthfulness-semantic-prompts}). These controls specify reasoning procedures, intermediate checks, or response organization. Truthfulness provides a broader standard for accepting and revising conclusions through justified claims, uncertainty handling, and correction of unsupported assertions. In Table~\ref{tab:truthfulness_semantic_ablation}, full truthfulness reaches 55.92\%, exceeding the controls by 2.23, 2.40, and 2.56 points and leading on every benchmark. Its advantage over verification supports our hypothesis that epistemic principles add value beyond explicit checking, providing a reusable behavioral standard that shapes teacher preferences throughout reasoning.

\begin{table}[h]
\centering
\setlength{\belowcaptionskip}{10pt}
\caption{\textbf{Fine-grained truthfulness PI ablations on Qwen3-1.7B.} Scores are $\mathrm{avg}@12$ accuracy (\%). Each replacement uses one checkpoint selected by the highest macro-average over ten checkpoints. Full truthfulness results are from Table~\ref{tab:math_main}. \textbf{Bold} denotes the best result in each column.}
\label{tab:truthfulness_semantic_ablation}
\small
\setlength{\tabcolsep}{8pt}
\renewcommand{\arraystretch}{1.10}
\begin{tabular}{lcccc}
\toprule
\textbf{PI variant} & \textbf{AMC23} & \textbf{AIME24} & \textbf{AIME25} & \textbf{Avg.} \\
\midrule
Generic reasoning & 76.91 & 49.17 & 35.00 & 53.69 \\
Verification & 77.51 & 47.50 & 35.56 & 53.52 \\
Presentation & 75.90 & 48.06 & 36.11 & 53.36 \\
\midrule
\textbf{Full truthfulness ($\tau$-OPSD)} & \textbf{78.31} & \textbf{51.39} & \textbf{38.06} & \textbf{55.92} \\
\bottomrule
\end{tabular}
\end{table}

\begin{figure}[htbp]
\centering
\includegraphics[width=\linewidth]{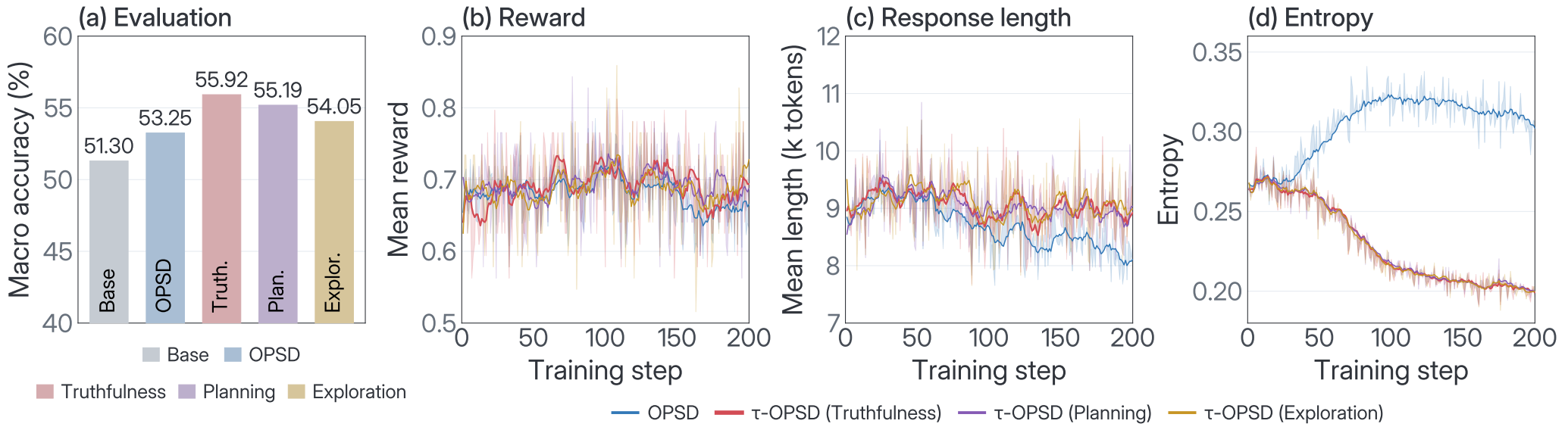}
\vspace{-20pt}
\caption{\textbf{Trait generalization at Qwen3-1.7B.} (a) Best-checkpoint macro-average accuracy on mathematical reasoning for Base, OPSD, and $\tau$-OPSD with truthfulness, planning, or exploration. (b–d) Training reward, response length, and entropy. Curves show 10-step moving averages. Shading shows deviations of raw values from these averages.}
\label{fig:trait_training}
\end{figure}

\paragraph{Generalization across Traits.}
To test generalization across behavioral traits, we replace the default truthfulness PI in $\tau$-OPSD with \textit{planning} or \textit{exploration} while keeping the remaining framework fixed. Planning emphasizes constraints, uncertainty, and revisable steps, whereas exploration encourages distinct reasoning paths followed by evidence-based convergence (prompts in Appendix~\ref{app:trait-pi}). On Qwen3-1.7B, planning and exploration achieve mathematical macro-average accuracies of 55.19\% and 54.05\%, respectively, compared with 51.30\% for Base and 55.92\% for truthfulness (Figure~\ref{fig:trait_training}(a)). Both also outperform solution-conditioned OPSD. These results support behavioral PI as a useful supervision source within $\tau$-OPSD beyond a single truthfulness instruction.

The three traits exhibit closely aligned training dynamics (Figure~\ref{fig:trait_training}(b--d)). All sustain training reward, keep response lengths near 9k tokens, and gradually reduce entropy from approximately 0.27 to 0.20. Exploration has slightly higher mean reward over the final 20 steps, though truthfulness achieves the best evaluation accuracy. Compared with OPSD's rising entropy and shorter responses, the trait variants maintain higher late-stage rewards with bounded response lengths. This shared pattern indicates that the framework accommodates different traits while preserving stable training. Appendix~\ref{app:truthfulness-scaling} provides further comparisons with OPSD and direct truthfulness substitution.

\section{Discussion and Conclusion}
\label{sec:discussion_conclusion}

\paragraph{Behavioral priors as reusable supervision.}
Our hypothesis locates the value of trait PI in reusable behavioral principles and the teacher preferences they induce. Truthfulness's advantage over procedural controls supports epistemic guidance for recurring reasoning decisions, while gains after removing the prompt at inference indicate that its influence can be internalized. Planning and exploration extend this perspective to other reasoning dispositions. The effectiveness of these principles also depends on how they are distilled. Direct truthfulness substitution in OPSD varies with scale (Appendix~\ref{app:truthfulness-scaling}), whereas stance gating improves learning by regulating the induced preferences against outcomes. Together, these findings link supervision content with training design, supporting behavioral alignment principles as a source of reasoning supervision when their influence is appropriately regulated.

\paragraph{Hard-example learning as supervision adaptation.}
Our findings suggest that difficult trajectories may require different supervision rather than simply more supervision. Correctness and resource-relative hardness capture complementary properties of a rollout, since incorrect trajectories are not necessarily difficult and budget-exhausted trajectories are not necessarily uninformative. $\tau$-OPSD therefore adapts both the privileged condition and the treatment of conflicting supervision according to trajectory state. This reframes hard-example learning from sample reweighting toward \emph{teaching adaptation}, where the form and reliability of guidance depend on the learner's state. More generally, behavioral priors should be used selectively when their induced preferences remain compatible with task outcomes, rather than distilled unconditionally.

\paragraph{Reasoning and abstention as complementary outcomes.}
Behavioral-prior training should be evaluated not only by whether a model solves more problems, but also by whether it responds appropriately when evidence is insufficient. The joint improvements in reasoning and abstention suggest that $\tau$-OPSD does not merely induce a more conservative policy, but may better distinguish when a conclusion is justified. We interpret this as evidence of improved evidence-sensitive answering rather than comprehensive honesty or proof that truthfulness itself causes the reasoning gains. Because behavioral traits may transfer differently across tasks, future work could explore traits such as corrigibility, fairness, harmlessness, and unbiasedness, and identify which priors are most useful for different domains and reasoning regimes.

\paragraph{Conclusion.}
We introduced $\tau$-OPSD, a trait-guided self-distillation framework that constructs reasoning supervision from shared behavioral instructions. It combines trait-conditioned guidance, outcome-aware stance gating, and adaptive supervision for hard samples. Across the evaluated settings, the method improves mathematical reasoning and shows favorable transfer to abstention behavior, while additional traits suggest that the framework extends beyond truthfulness. Overall, effective self-teaching depends not only on teacher information, but also on the behavioral prior and how its supervision is adapted to the learner's state. This points to behavioral traits as general-purpose training signals rather than solely alignment targets.

\clearpage
\setcounter{tocdepth}{3}
\tableofcontents

\newpage
\appendix
\section{Appendix}

\subsection{Training Algorithm}
\label{app:tau-algorithm}

Algorithm~\ref{alg:tau-opsd} summarizes trajectory collection, guidance selection, gated distillation, and teacher updates. Here, $p$ denotes the student distribution, while $q^z$ and $q^0$ denote the shared teacher with and without PI. The compact versions of $p$ and $q^z$ are $P$ and $Q$. The operator $\operatorname{sg}$ stops gradients, and $r_{i,t}$ is a detached rollout-engine correction (default $1$). Operations indexed by $i$ apply to every trajectory, and token sums exclude padding. Default settings follow Section~\ref{sec:tau-method}.

\begin{algorithm}[H]
\caption{$\tau$-OPSD: trait-guided on-policy self-distillation}
\label{alg:tau-opsd}
\small
\begin{algorithmic}[1]
\Statex \hspace*{-\leftmargin}\textbf{Input:} Student $\theta$; data $\mathcal D$; binary verifier $V$; truthfulness and local-progress PI $z_{\mathrm{truth}},z_{\mathrm{local}}$.
\Statex \hspace*{-\leftmargin}\textbf{Require:} Batch size $B$; token budget $L_{\max}$; gate scale $\rho$; EMA decay $\mu$; stabilizer $\epsilon$; top-$K$ count $K$.
\State $\phi\gets\theta$ \Comment{Initialize teacher parameters}
\For{each training iteration}
    \State Sample $\{x_i\}_{i=1}^{B}\subset\mathcal D$; rollout $y_i\sim p_\theta(\cdot\mid x_i)$ up to $L_{\max}$ tokens.
    \State $T_i\gets |y_i|,\quad R_i\gets V(x_i,y_i),\quad H_i\gets\mathbf{1}[T_i\ge L_{\max}]$ \Comment{Length, correctness, hard flag}
    \State $z_i\gets\begin{cases}z_{\mathrm{truth}},&H_i=0,\\z_{\mathrm{local}},&H_i=1.\end{cases}$
    \State Score $p_{i,t},q^{z_i}_{i,t},q^0_{i,t}$ on the same student prefixes. \Comment{Eq.~\ref{eq:tau-policies}}
    \State $s_i\gets\displaystyle\frac1{T_i}\sum_{t=1}^{T_i}\log\frac{q^{z_i}_{i,t}(y_{i,t})}{q^0_{i,t}(y_{i,t})}$ \Comment{PI-induced stance, Eq.~\ref{eq:tau-stance}}
    \State $c_i\gets(2R_i-1)s_i$ \Comment{Outcome consistency}
    \State $w_i\gets\begin{cases}c_i,&c_i\ge0,\\\rho c_i,&c_i<0\ \text{and}\ H_i=0,\\0,&c_i<0\ \text{and}\ H_i=1.\end{cases}$ \Comment{Stance gate}
    \State $\widehat w_i\gets\operatorname{sg}\!\left[w_i\big/\left(B^{-1}\sum_{j=1}^{B}|w_j|+\epsilon\right)\right]$
    \State Construct $P_{i,t},Q_{i,t}$ on the PI teacher's top-$K$ tokens plus one tail bucket.
    \State $\mathcal L\gets\displaystyle\frac1B\sum_{i=1}^{B}\widehat w_i\frac1{T_i}\sum_{t=1}^{T_i}r_{i,t}\,\mathrm{JSD}_{1/2}(P_{i,t},Q_{i,t})$ \Comment{Eqs.~\ref{eq:tau-jsd},~\ref{eq:tau-loss}}
    \State $\theta\gets\operatorname{OptimizerStep}(\theta,\nabla_\theta\mathcal L)$
    \State $\phi\gets\mu\phi+(1-\mu)\theta$ \Comment{EMA teacher}
\EndFor
\State \Return $p_\theta$
\end{algorithmic}
\end{algorithm}

\paragraph{Trajectory scoring and gating.}
The student samples one response per question without PI. Budget exhaustion selects local-progress guidance independently of verifier correctness (Eq.~\ref{eq:tau-routing}). The two teacher branches share weights and rescore the same student prefixes; no additional teacher trajectory is generated. Averaging the token log-likelihood ratios gives stance $s_i$, which is combined with verifier correctness as $c_i=(2R_i-1)s_i$ to measure outcome consistency. The gate then assigns attractive, weakly suppressive, or zero supervision, with batch normalization preserving the sign (Eqs.~\ref{eq:tau-gate}--\ref{eq:tau-weight-norm}).

\paragraph{Distribution matching.}
\label{app:tau-implementation}
For the loss, $P_{i,t}$ and $Q_{i,t}$ use the same $K=100$ tokens selected by the PI teacher. Each retains the original probabilities on these tokens and places all remaining mass in one tail bucket, yielding $K+1$ entries without renormalizing the top-$K$ probabilities. The stance calculation uses observed-token likelihoods, while JSD matches these compact distributions (Eq.~\ref{eq:tau-jsd}). The detached rollout-engine correction $r_{i,t}$ defaults to $1$ when engine discrepancies are ignored.

\paragraph{Gradient flow and update order.}
Teacher predictions, trajectory weights, and rollout-engine corrections are held fixed during the student update. Gradients flow through the student distribution, including its contribution to the JSD mixture. Token losses are averaged within each trajectory before applying its signed weight and averaging across the batch (Eq.~\ref{eq:tau-loss}). The EMA update follows the student optimizer step (Eq.~\ref{eq:tau-ema}); only the resulting student is needed at inference.

\subsection{Experimental Setup}
\label{app:experimental_setup}

This section supplements the main experimental setup with dataset composition, baseline implementations, and evaluation protocols. In particular, we detail how the abstention evaluation distinguishes missing information from an incorrect answer to a fully specified question.

\subsubsection{Datasets}
\label{app:datasets}

\paragraph{DeepMath-103K.}
DeepMath-103K~\citep{he2025deepmath} supplies training problems with difficulty annotations, verifiable final answers, and reference solutions. We use the complete preprocessed subsets within the model-specific difficulty ranges reported in Section~\ref{sec:exp-setup}; fixed-step training need not traverse every question. Final answers support outcome verification for all training methods. Access to reference solutions in teacher contexts depends on the method, as specified below.

\paragraph{AMC23.}
We use the AIMO AMC validation collection~\citep{aimoAMC}, containing 83 competition problems from 2022--2023. The label AMC23 denotes this fixed collection throughout the paper. Each problem is evaluated against its reference answer using the mathematical verification pipeline.

\paragraph{AIME24.}
We use the 30 problems from the 2024 portion of the AIMO AIME validation collection~\citep{aimoAIME}. This benchmark evaluates competition-level mathematical reasoning, with final answers checked against the supplied references.

\paragraph{AIME25.}
We use the OpenCompass AIME2025 release~\citep{opencompassAIME2025}, containing 30 problems from the two 2025 examination rounds. It is evaluated separately from AIME24 under the same sampling and scoring protocol.

\paragraph{AbstentionBench.}
We evaluate three underspecified-reasoning subsets of AbstentionBench~\citep{kirichenko2025abstentionbench}. Its paired construction associates an original, answerable question with a version whose preceding context is removed, retaining the final question and any answer choices. This tests whether a model recognizes insufficient information while continuing to answer fully specified questions. We follow the official filtering and context-removal procedure and retain its answerability labels without manual relabeling. These subsets are used only for evaluation, measuring transfer from mathematical training to evidence-sensitive abstention.

\paragraph{GSM8K-Abstain.}
This subset is derived from the GSM8K test split~\citep{cobbe2021gsm8k}, which contains mathematical word problems. The official filter retains 1,213 source questions, yielding 1,213 answerable--unanswerable pairs, or 2,426 evaluated questions. It probes whether the model detects missing context needed for a numerical answer.

\paragraph{MMLU-Math-Abstain.}
This subset uses three MMLU test subjects~\citep{hendrycks2021mmlu}: college mathematics, abstract algebra, and high-school mathematics. They contribute 32, 1, and 100 retained pairs, respectively, for a total of 133 pairs and 266 questions. Answer choices are preserved when context is removed, so selecting an available option alone does not establish that the question is answerable.

\paragraph{GPQA-Abstain.}
This subset is constructed from GPQA Diamond~\citep{rein2023gpqa}. The official filter retains 40 of its 198 source questions, producing 40 pairs and 80 evaluated questions. We freeze the choice ordering during preprocessing and use the original answer text as the judging reference. Together, the three abstention subsets contain 1,386 pairs, with equal numbers of answerable and unanswerable questions within each subset.

\subsubsection{Baselines}
\label{app:baselines}

\paragraph{Base Model.}
Base denotes the original released Qwen3 checkpoint~\citep{yang2025qwen3} before our additional training. It provides the reference for measuring gains from post-training at each model scale. This label does not refer to the pretrained-only \texttt{-Base} model series.

\paragraph{GRPO.}
Group Relative Policy Optimization~\citep{shao2024deepseekmath} uses verifiable outcomes to estimate relative advantages within a group of sampled responses. It updates the policy without a separately trained critic. This baseline represents outcome-driven reinforcement learning, with supervision determined by the rewards of sampled solutions.

\paragraph{OPSD.}
On-Policy Self-Distillation~\citep{zhao2026opsd} conditions a self-teacher on reference solutions and answers, then transfers its token-level predictions on student-generated prefixes. Our OPSD baseline keeps the teacher weights fixed during training. It provides the comparison with instance-specific, solution-privileged supervision; $\tau$-OPSD instead conditions its EMA teacher on behavioral instructions.

\paragraph{SDPO.}
Self-Distillation Policy Optimization~\citep{hubotter2026sdpo} converts textual feedback into dense self-distillation supervision. Our implementation follows the official construction using successful sibling rollouts of the same question, excluding the trajectory currently being scored. These sampled peer solutions form the privileged teacher context. We do not substitute dataset reference solutions for sibling feedback, preserving the distinction between SDPO and the solution-conditioned OPSD baseline.

\subsubsection{Evaluation Protocol}
\label{app:evaluation-protocol}

\paragraph{Mathematical reasoning.}
For each question, we generate 12 responses in thinking mode with temperature 0.6, top-$p=0.95$, top-$k=20$, and a maximum completion length of 38,912 tokens. Let $a_{ij}\in\{0,1\}$ denote the verified correctness of response $j$ to question $i$. For a benchmark with $N$ questions, we report
\begin{equation}
    \mathrm{avg}@k=\frac{1}{Nk}\sum_{i=1}^{N}\sum_{j=1}^{k}a_{ij},
    \qquad k=12.
    \label{eq:app-avgk}
\end{equation}
This averages individual response accuracy, rather than majority-vote accuracy or the probability that at least one response is correct. The overall mathematical score is the unweighted mean across AMC23, AIME24, and AIME25. This macro-average determines checkpoint selection: all three benchmark scores are reported from the same selected checkpoint.

\paragraph{Abstention generation.}
Each question receives one response with thinking disabled, temperature 0.8, top-$p=0.95$, no top-$k$ truncation, and a 4,000-token completion limit. The evaluated model receives the question and any answer choices; reference answers and answerability labels are withheld. The paired inputs are evaluated individually, so the model does not receive the original question as additional context when answering its underspecified counterpart.

\paragraph{Abstention and correctness judging.}
We retain the official AbstentionBench judging prompts~\citep{kirichenko2025abstentionbench}, substituting a fixed Qwen3-8B judge for the original benchmark judge. The judge uses nonthinking greedy decoding with at most 100 output tokens per judgment. It assesses abstention on every response and separately assesses correctness on answerable questions. Judgments use the final response, with reference answers and answerability labels supplied only to the judge as required by the corresponding prompt. Empty responses are retained and judged through the same procedure rather than discarded or automatically counted as abstentions.

\paragraph{Abstention metrics.}
We treat unanswerable questions as the positive class. A true positive ($TP$) is an abstention on an unanswerable question; a false positive ($FP$) is an abstention on an answerable question; and a false negative ($FN$) is a non-abstaining response to an unanswerable question. Precision, recall, and F1 are
\begin{equation}
    P=\frac{TP}{TP+FP},\qquad
    R=\frac{TP}{TP+FN},\qquad
    F_1=\frac{2TP}{2TP+FP+FN},
    \label{eq:app-abstention}
\end{equation}
with undefined values set to zero. Answerable accuracy is the fraction of \emph{all} answerable questions whose responses are judged correct; abstentions remain in its denominator. Reporting both metrics reveals whether improved abstention comes at the cost of answering well-specified questions. Scores are expressed as percentages, and each overall score is the unweighted mean of the three subset scores, so the larger GSM8K-Abstain subset does not dominate the aggregate.

\subsubsection{Hyperparameters}
\label{app:hyperparameters}

\paragraph{Hardware and implementation.}
Training uses Ascend 910C NPUs (approximately 65\,GB each): 16 on one node for Qwen3-1.7B and 32 across two nodes for Qwen3-4B/8B. We use FSDP2, vLLM-Ascend, BF16, gradient checkpointing, and CPU offloading of parameters and optimizer states. All methods process 64 questions per training step in the same unshuffled order.

\begin{table}[htbp]
\centering
\setlength{\belowcaptionskip}{10pt}
\caption{\textbf{Shared training settings.} Length limits apply to student inputs and responses.}
\label{tab:hyperparameters_shared}
\small
\renewcommand{\arraystretch}{1.10}
\setlength{\tabcolsep}{8pt}
\begin{tabular}{ll}
\toprule
\textbf{Parameter} & \textbf{Setting} \\
\midrule
Training steps / checkpoint interval & 200 / 20 \\
Global question batch size & 64 \\
Update epochs per rollout batch & 1 \\
Optimizer / weight decay & AdamW / 0.01 \\
Learning rate & $1\times10^{-6}$ \\
Learning-rate schedule & 50-step linear warmup, then constant \\
Maximum prompt / completion length & 2,048 / 16,384 tokens \\
Rollout temperature / top-$p$ / top-$k$ & 1.0 / 0.95 / 20 \\
Rollout importance correction & Token-level, capped at 2.0 \\
\bottomrule
\end{tabular}
\end{table}

\begin{table}[htbp]
\centering
\setlength{\belowcaptionskip}{10pt}
\caption{\textbf{Method-specific settings.} Mini-batches use question-equivalent units. Distillation uses $K=100$ tokens plus a tail bucket.}
\label{tab:hyperparameters_methods}
\small
\renewcommand{\arraystretch}{1.10}
\setlength{\tabcolsep}{4pt}
\begin{tabular}{lcccc}
\toprule
\textbf{Parameter} & \textbf{GRPO} & \textbf{OPSD} & \textbf{SDPO} & \textbf{$\tau$-OPSD} \\
\midrule
Rollouts per question & 8 & 8 & 8 & 1 \\
Mini-batch size (questions) & 16 & 16 & 16 & 64 \\
Optimizer updates per training step & 4 & 4 & 4 & 1 \\
Gradient clipping norm & 1.0 & 0.1 & 1.0 & 0.1 \\
Loss aggregation & Token mean & Token mean & \shortstack{Active-token\\mean} & \shortstack{Sequence mean\\of token means} \\
\midrule
Objective & \shortstack{Clipped policy\\gradient} & Forward KL & JSD & JSD \\
JSD mixture coefficient & -- & -- & 0.5 & 0.5 \\
KL-to-reference coefficient & $10^{-3}$ & 0 & 0 & 0 \\
PPO ratio clipping range & $[0.8,\,1.2]$ & -- & -- & -- \\
Divergence upper clip & -- & 0.05 & None & None \\
Top-$K$ selection policy & -- & Teacher & Student & Teacher \\
Teacher / EMA decay & -- & Fixed & EMA / 0.95 & EMA / 0.99 \\
Additional current/old-policy IS cap & -- & -- & 2.0 & -- \\
\bottomrule
\end{tabular}
\end{table}

\paragraph{Configuration rationale.}
GRPO needs multiple responses for group-relative advantages, and SDPO uses sibling feedback. OPSD and $\tau$-OPSD permit single rollouts, although our OPSD baseline uses eight. Comparisons evaluate complete configurations, with component effects examined through ablations (Appendix~\ref{app:full_ablation}).

\paragraph{Teacher and weighting settings.}
SDPO updates its EMA teacher per training step, and $\tau$-OPSD per optimizer step (Eq.~\ref{eq:tau-ema}). In $\tau$-OPSD, outcome-inconsistent trajectories ($c_i<0$) receive weights of $0.1c_i$ for ordinary samples and zero for hard samples. Absolute-value normalization preserves their signs (Section~\ref{sec:tau-gating}).

\paragraph{Checkpoint evaluation.}
We evaluate every saved checkpoint with 12 responses per question on AMC23, AIME24, and AIME25. For each run, all mathematical scores come from the single checkpoint with the highest unweighted mean of the three $\mathrm{avg}@12$ scores (Appendix~\ref{app:evaluation-protocol}).

\subsection{Extended Experiment Analysis}

\subsubsection{Full Component Ablation}
\label{app:full_ablation}

Table~\ref{tab:ablation_full} reports component ablations on Qwen3-1.7B. \textit{W/o JSD} uses forward KL, as in OPSD, and \textit{w/o EMA teacher} fixes the teacher throughout training. \textit{W/o stance gate} gives all samples equal weight. \textit{W/o truthfulness} replaces truthfulness PI with the full reference derivation and final answer, while \textit{w/o hard sample} treats all trajectories as ordinary, using truthfulness PI and the ordinary-trajectory gate even when the token budget is exhausted. Other settings remain unchanged.

Beyond the core components, replacing JSD or fixing the teacher reduces macro-average accuracy by 3.44 and 2.90 points, respectively. The former supports symmetric distribution matching in this setting, while the latter highlights the value of keeping teacher supervision aligned with the evolving student. Both choices provide effective optimization support for trait-guided supervision.

\begin{table}[h]
\centering
\setlength{\belowcaptionskip}{10pt}
\caption{\textbf{Full component ablation on Qwen3-1.7B for mathematical reasoning.}}
\label{tab:ablation_full}

\small
\setlength{\tabcolsep}{6pt}
\renewcommand{\arraystretch}{1.10}

\begin{tabular}{lcccc}
\toprule
\textbf{Ablation}
& \textbf{AMC23}
& \textbf{AIME24}
& \textbf{AIME25}
& \textbf{Avg.} \\
\midrule

Base
& 75.60
& 44.70
& 33.60
& 51.30 \\

w/o JSD
& 75.20
& 46.67
& 35.56
& 52.48 \\

w/o EMA teacher
& 76.00
& 47.78
& 35.28
& 53.02 \\

w/o stance gate
& 75.50
& 46.94
& 34.44
& 52.29 \\

w/o truthfulness
& \underline{78.11}
& 49.44
& 35.56
& 54.37 \\

w/o hard sample
& 76.31
& \textbf{51.94}
& \underline{35.83}
& \underline{54.69} \\

\textbf{$\tau$-OPSD (ours)}
& \textbf{78.31}
& \underline{51.39}
& \textbf{38.06}
& \textbf{55.92} \\

\bottomrule
\end{tabular}
\end{table}

\paragraph{Disentangling Stance and Outcome Weighting.}
On Qwen3-1.7B, we replace stance gating with two outcome-based rules, retaining PI routing and other training settings. Before absolute-mean normalization, \emph{Correct-only} weights correct trajectories by $+1$ and all others by zero. \emph{Outcome-only} additionally weights incorrect ordinary trajectories by $-0.1$, keeping incorrect budget-exhausted trajectories at zero. Both rules omit stance scores. In Table~\ref{tab:stance_outcome_ablation}, Correct-only gains 0.74 points over uniform weighting, and Outcome-only adds 1.46 points. Full $\tau$-OPSD gains a further 1.43 points and leads on all three benchmarks, supporting the complementary value of outcome-based weighting and trait-induced stance beyond correctness alone.

\begin{table}[htbp]
\centering
\setlength{\belowcaptionskip}{10pt}
\caption{\textbf{Disentangling stance and outcome weighting on Qwen3-1.7B.} Scores are $\mathrm{avg}@12$ accuracy (\%) from the checkpoint selected by mathematical macro-average. \textbf{Bold} denotes the best result in each column.}
\label{tab:stance_outcome_ablation}
\small
\setlength{\tabcolsep}{8pt}
\renewcommand{\arraystretch}{1.10}
\begin{tabular}{lcccc}
\toprule
\textbf{Weighting rule} & \textbf{AMC23} & \textbf{AIME24} & \textbf{AIME25} & \textbf{Avg.} \\
\midrule
Uniform (w/o stance gate) & 75.50 & 46.94 & 34.44 & 52.29 \\
Correct-only & 75.20 & 48.33 & 35.56 & 53.03 \\
Outcome-only & 76.80 & 49.17 & 37.50 & 54.49 \\
\midrule
\textbf{$\tau$-OPSD (ours)} & \textbf{78.31} & \textbf{51.39} & \textbf{38.06} & \textbf{55.92} \\
\bottomrule
\end{tabular}
\end{table}

\subsubsection{Training Dynamic Analysis}

\begin{figure}[h]
\centering
\includegraphics[width=\linewidth]{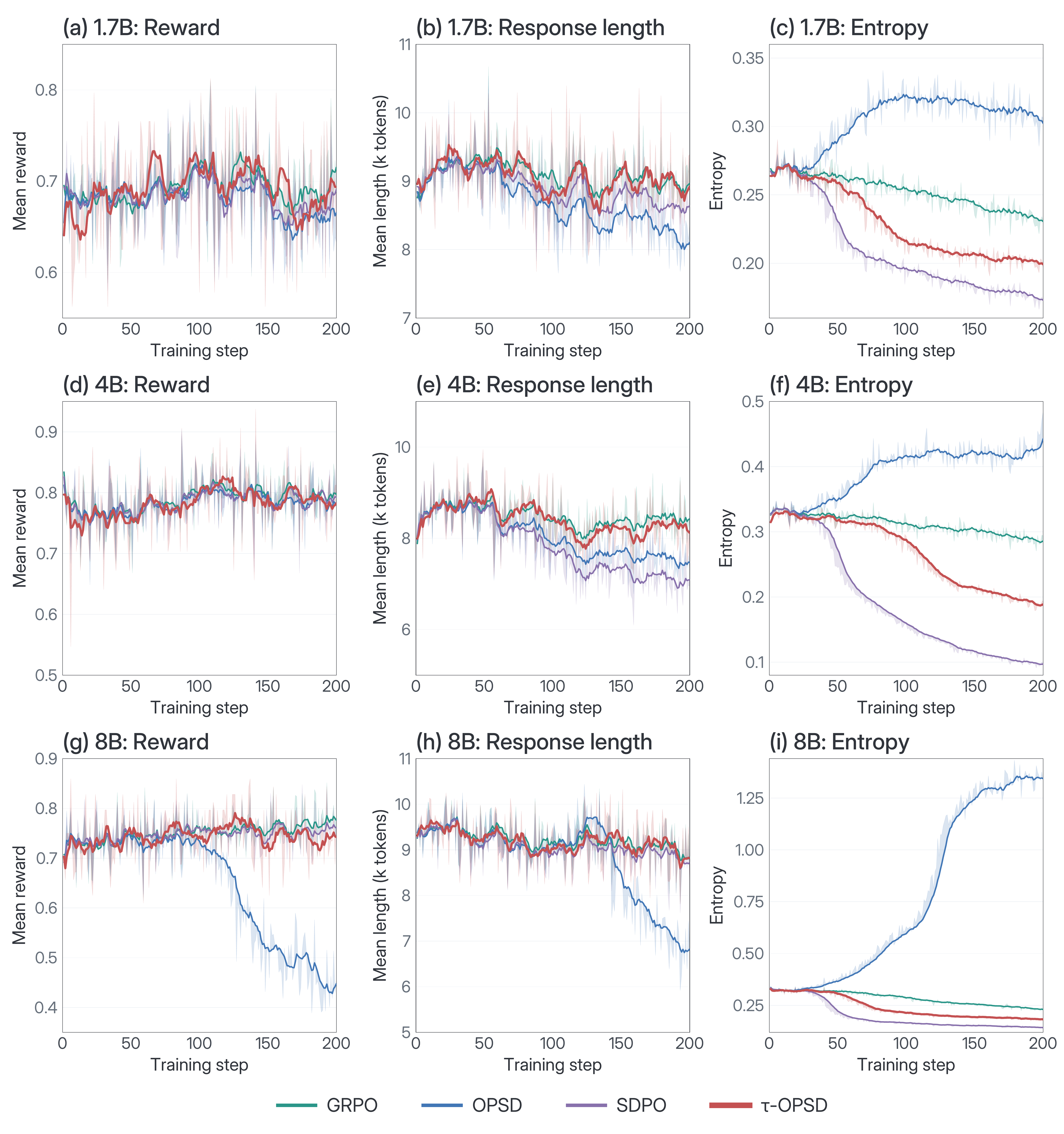}
\caption{\textbf{Training dynamics across model scales.} Reward, response length, and policy entropy for $\tau$-OPSD, GRPO, OPSD, and SDPO at 1.7B, 4B, and 8B (top to bottom). Solid curves show 10-step moving averages; shaded regions indicate deviations of raw values from these averages. $\tau$-OPSD is highlighted with thicker red curves. Vertical scales vary across panels.}
\label{fig:training_dynamic_multiscale}
\end{figure}

Training Dynamics. The Figure ~\ref{fig:training_dynamic_multiscale} compares $\tau$-OPSD with GRPO, OPSD, and SDPO across three model scales. $\tau$-OPSD maintains relatively stable training rewards and response lengths while gradually reducing policy entropy. It avoids the pronounced degradation observed for OPSD at 8B and retains higher entropy than SDPO, suggesting a balance between policy concentration and distributional diversity.

Compared with OPSD, $\tau$-OPSD maintains more stable training. OPSD's entropy rises at all scales, accompanied at 8B by declining reward and shorter responses. Its mean reward over the final 20 steps is 0.446, versus 0.738 for $\tau$-OPSD. This joint deterioration suggests policy degradation rather than improved reasoning efficiency from shorter outputs.

Compared with SDPO, $\tau$-OPSD retains higher entropy while keeping response lengths bounded or gradually decreasing. At 4B, mean entropy over the final 20 steps is 0.191 for $\tau$-OPSD versus 0.099 for SDPO. This combination preserves distributional diversity while keeping response lengths controlled.

Compared with GRPO, $\tau$-OPSD achieves competitive training rewards with a greater entropy reduction, although GRPO generally has slightly higher late-stage rewards. Its entropy lies between GRPO's higher values and SDPO's stronger concentration. Together with the evaluation gains, these dynamics highlight stable training with controlled policy concentration.

\subsubsection{Training-Time versus Inference-Time Trait Conditioning}
\label{app:trait-conditioning}

Table~\ref{tab:icl_comparison} complements the aggregate results in Figure~\ref{fig:sensitivity_generalization}(a) with a per-benchmark breakdown. Here, \emph{in-context learning (ICL)} refers to conditioning on the truthfulness instruction at inference without parameter updates. ICL applies this instruction to Base, while $\tau$-OPSD+ICL adds it to the same checkpoint selected for $\tau$-OPSD, keeping checkpoint selection fixed. Without inference-time trait prompting, $\tau$-OPSD exceeds ICL on all nine model--benchmark combinations, with macro-average gains of 3.28, 2.23, and 1.96 percentage points at 1.7B, 4B, and 8B, respectively.

The breakdown also reveals differences obscured by the average. Reintroducing the instruction after distillation lowers accuracy on all three benchmarks at 1.7B and 4B. At 8B, it improves AMC23 and AIME24 by 1.21 and 0.83 points but reduces AIME25 by 1.67 points, leaving only a 0.13-point average gain. Thus, the benefits of trait-guided training persist without the trait prompt at inference, while the effect of additional prompting varies across model scales and benchmarks.

\begin{table}[h]
\centering
\setlength{\belowcaptionskip}{10pt}
\caption{
\textbf{Training-time versus inference-time trait conditioning.}
ICL (in-context learning) adds the truthfulness instruction at inference.
Within each model scale and metric, \textbf{bold} and
\underline{underlining} indicate the best and second-best results, respectively.
}
\label{tab:icl_comparison}

\small
\setlength{\tabcolsep}{5.5pt}
\renewcommand{\arraystretch}{1.10}

\begin{tabular}{llcccc}
\toprule
\textbf{Model}
& \textbf{Method}
& \textbf{AMC23}
& \textbf{AIME24}
& \textbf{AIME25}
& \textbf{Avg. Acc.} \\
\midrule

\multirow{4}{*}{Qwen3-1.7B}
& Base
& 75.60
& 44.70
& 33.60
& 51.30 \\

& ICL
& \underline{75.70}
& 45.28
& \underline{36.94}
& 52.64 \\

& $\tau$-OPSD
& \textbf{78.31}
& \textbf{51.39}
& \textbf{38.06}
& \textbf{55.92} \\

& $\tau$-OPSD+ICL
& 74.60
& \underline{48.89}
& 36.67
& \underline{53.39} \\

\midrule

\multirow{4}{*}{Qwen3-4B}
& Base
& 87.75
& 70.00
& 64.17
& 73.97 \\

& ICL
& 88.95
& 71.39
& \underline{65.28}
& 75.21 \\

& $\tau$-OPSD
& \textbf{89.26}
& \textbf{75.56}
& \textbf{67.50}
& \textbf{77.44} \\

& $\tau$-OPSD+ICL
& \underline{88.96}
& \underline{75.00}
& 64.17
& \underline{76.04} \\

\midrule

\multirow{4}{*}{Qwen3-8B}
& Base
& 89.26
& 72.50
& 68.30
& 76.69 \\

& ICL
& 89.85
& 74.44
& 68.61
& 77.63 \\

& $\tau$-OPSD
& \underline{90.16}
& \underline{78.06}
& \textbf{70.56}
& \underline{79.59} \\

& $\tau$-OPSD+ICL
& \textbf{91.37}
& \textbf{78.89}
& \underline{68.89}
& \textbf{79.72} \\

\bottomrule
\end{tabular}
\end{table}

\subsubsection{Training Efficiency}
\label{app:training-efficiency}

Table~\ref{tab:training_step_time} reports average total training-step time under the configurations in Section~\ref{sec:exp-setup}. Across the three model scales, $\tau$-OPSD achieves per-step speedups of $1.94$--$2.66\times$ over OPSD, $1.66$--$2.23\times$ over SDPO, and $1.56$--$1.94\times$ over GRPO. A key source of savings is the reduced rollout budget: each batch contains 64 questions, but $\tau$-OPSD samples only one response per question, compared with eight for the baselines. Trait-conditioned supervision scores the existing student trajectory without generating an additional teacher solution. Fewer sampled trajectories therefore reduce both generation and subsequent trajectory-processing costs. These timings measure per-step cost under the reported configurations, including their response lengths, teacher scoring, and update schedules.

\begin{table}[htbp]
\centering
\setlength{\belowcaptionskip}{10pt}
\caption{\textbf{Average total training-step time (seconds).} Each step uses 64 questions, with one rollout per question for $\tau$-OPSD and eight for the baselines. \textbf{Bold} indicates the lowest time at each scale.}
\label{tab:training_step_time}
\small
\setlength{\tabcolsep}{12pt}
\renewcommand{\arraystretch}{1.10}
\begin{tabular}{lrrrr}
\toprule
\textbf{Model size} & \textbf{GRPO} & \textbf{OPSD} & \textbf{SDPO} & \textbf{$\tau$-OPSD} \\
\midrule
1.7B & 532.7 & 731.4 & 613.6 & \textbf{275.0} \\
4B   & 482.0 & 599.3 & 513.8 & \textbf{308.8} \\
8B   & 587.2 & 727.6 & 724.8 & \textbf{350.6} \\
\bottomrule
\end{tabular}
\end{table}

\paragraph{Controlled baseline comparisons.}
Table~\ref{tab:controlled_baselines} separates supervision gains from rollout and optimization choices on Qwen3-1.7B. Single-update SDPO uses a mini-batch of 64 and one optimizer update per training step, retaining eight rollouts and its original objective. OPSD first reduces rollouts to one, then uses a mini-batch of 64 and one update. Recipe-aligned OPSD additionally adopts EMA teacher updates, JSD without contribution clipping, and sequence-level normalization, retaining solution PI. Single-rollout OPSD has per-step time comparable to $\tau$-OPSD, clarifying that the speedups above reflect the original rollout configurations. Nevertheless, $\tau$-OPSD exceeds the strongest OPSD control by 2.20 points, recipe-aligned OPSD by 2.60 points, and single-update SDPO by 3.37 points. These comparisons support gains beyond rollout counts, update schedules, and general optimization choices.

\begin{table}[htbp]
\centering
\setlength{\belowcaptionskip}{10pt}
\caption{\textbf{Controlled baseline comparisons on Qwen3-1.7B.} Scores are $\mathrm{avg}@12$ accuracy (\%). Rollouts are per question and updates are per training step. \textbf{Bold} indicates the best result in each metric.}
\label{tab:controlled_baselines}
\small
\setlength{\tabcolsep}{4pt}
\renewcommand{\arraystretch}{1.10}
\begin{tabular}{lrrcccc}
\toprule
\textbf{Method} & \textbf{Rollouts} & \textbf{Updates} & \textbf{AMC23} & \textbf{AIME24} & \textbf{AIME25} & \textbf{Avg.} \\
\midrule
SDPO (single-update) & 8 & 1 & 74.30 & 48.06 & 35.28 & 52.55 \\
OPSD (single-rollout) & 1 & 4 & 75.80 & 48.89 & 34.44 & 53.05 \\
OPSD (single-update) & 1 & 1 & 75.90 & 49.44 & 35.83 & 53.72 \\
OPSD (recipe-aligned) & 1 & 1 & 75.50 & 50.28 & 34.17 & 53.32 \\
\midrule
$\tau$-OPSD (ours) & 1 & 1 & \textbf{78.31} & \textbf{51.39} & \textbf{38.06} & \textbf{55.92} \\
\bottomrule
\end{tabular}
\end{table}

\subsubsection{Hard-Sample Learning and Stance Gating}
\label{app:hard-gate-mechanisms}

On Qwen3-1.7B, removing stance gating or hard-sample learning lowers mathematical macro-average accuracy by 3.63 and 1.23 points, respectively (Table~\ref{tab:ablation_full}). We use Qwen3-8B to illustrate both mechanisms over 200 training steps, followed by a controlled one-step diagnostic on Qwen3-1.7B. Interval statistics use raw observations.

\begin{figure}[htbp]
\centering
\includegraphics[width=\linewidth]{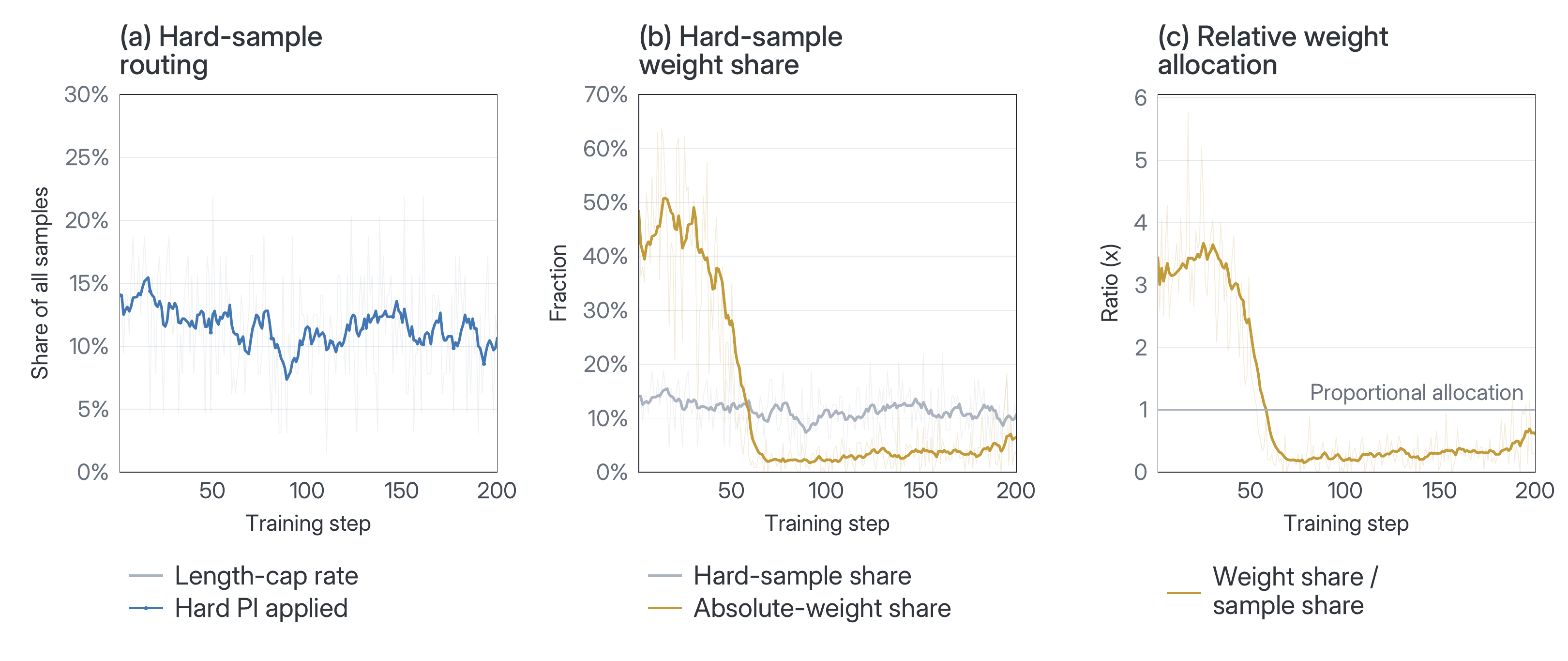}
\vspace{-10pt}
\caption{\textbf{Hard-sample dynamics on Qwen3-8B.} (a) Length-cap and hard-PI rates. (b) Sample and absolute-weight shares. (c) Absolute-weight share divided by sample share. Faint lines show raw values. Thick lines show trailing ten-step means, with (c) using the ratio of window-averaged shares.}
\label{fig:hard_sample_mechanism}
\end{figure}

\paragraph{Adaptive hard-sample supervision.}
Budget-exhausted trajectories receive local-progress PI, while outcome-inconsistent guidance is masked (Eqs.~\ref{eq:tau-routing} and~\ref{eq:tau-gate}). Figure~\ref{fig:hard_sample_mechanism} shows matching length-cap and hard-PI activation rates but changing weight allocation. During steps 1--40, hard samples account for 13.05\% of trajectories and 43.06\% of absolute weight. During steps 61--100, these shares fall to 10.39\% and 2.16\%. Relative allocation, defined as absolute-weight share divided by sample share, thus decreases from $3.30\times$ to $0.21\times$. Hard-sample supervision remains active, with its share jointly regulated by stance magnitude and gating as well as sample prevalence.

\begin{figure}[H]
\centering
\includegraphics[width=\linewidth]{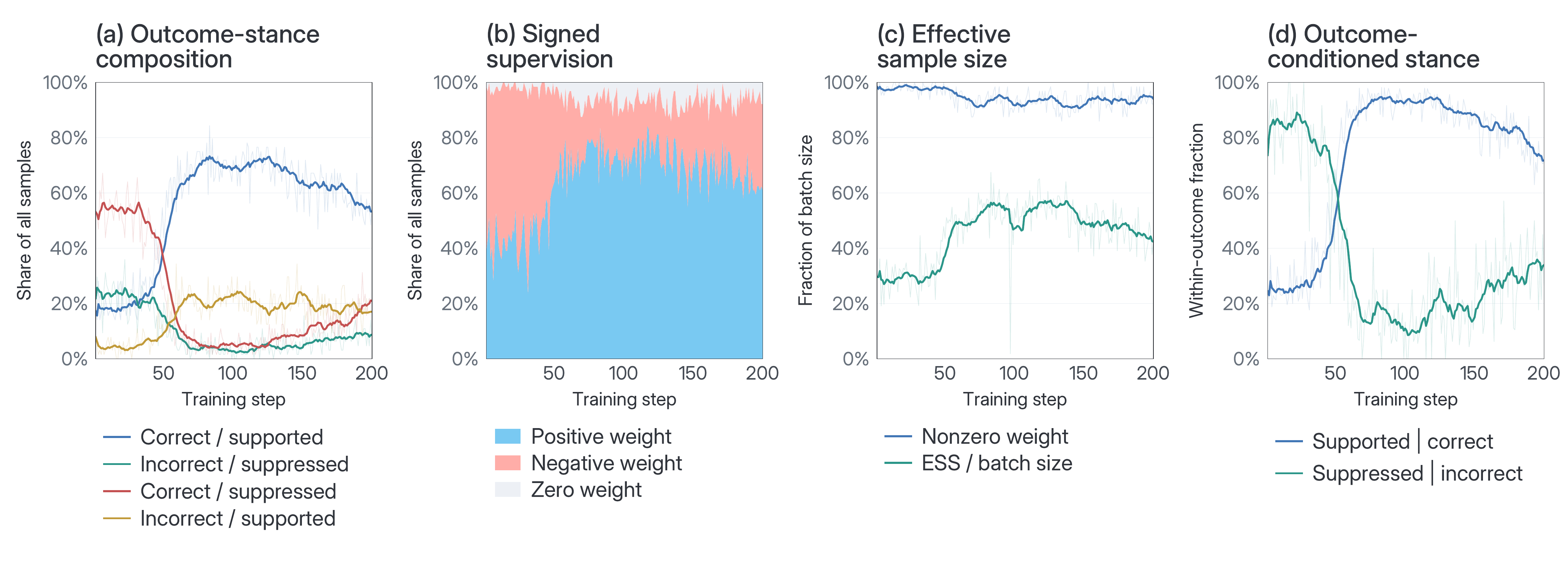}
\vspace{-10pt}
\caption{\textbf{Stance-gate dynamics on Qwen3-8B.} (a) Outcome--stance composition. (b) Signed-weight proportions. (c) Nonzero-weight fraction and normalized ESS. (d) Outcome-conditioned stance rates. Faint lines show raw values and thick lines use trailing ten-step windows. Conditional rates pool window counts. Stacked proportions are unsmoothed.}
\label{fig:stance_gate_mechanism}
\end{figure}

\paragraph{Outcome-aware supervision.}
On average, 94.45\% of samples retain nonzero supervision weights, while the negative-weight fraction decreases from 55.35\% in steps 1--40 to 20.00\% in steps 61--100 (Figure~\ref{fig:stance_gate_mechanism}). The gate therefore adjusts supervision direction while retaining most samples. Comparing these intervals, PI increases teacher likelihood for a larger fraction of correct trajectories and decreases it for a smaller fraction of incorrect trajectories, relative to the teacher without PI (Figure~\ref{fig:stance_gate_mechanism}(d)). The gate combines these PI-induced likelihood preferences with verifier outcomes through $c_i=(2R_i-1)s_i$, assigning positive, weakly negative, or zero distillation weights (Eq.~\ref{eq:tau-gate}). This outcome-aware weighting complements trait guidance, consistent with its ablation advantage over uniform weighting.

\paragraph{Weight distribution.}
Normalized effective sample size measures how broadly absolute weights are distributed:
\begin{equation}
    \frac{\mathrm{ESS}}{B}=\frac{\bigl(\sum_i|w_i|\bigr)^2}{B\sum_i w_i^2}.
    \label{eq:app-weight-ess}
\end{equation}
Its mean rises from 29.31\% in steps 1--40 to 53.24\% in steps 101--160, indicating broader supervision despite transient concentration, including the minimum at step 97. Together, the curves show complementary adaptation of the teaching condition, supervision direction, and weight allocation.

\paragraph{Paired one-step gate intervention.}
From a Qwen3-1.7B step-100 checkpoint, we collect four batches of 64 trajectories, with one rollout per question. Each batch undergoes paired Full and Positive-only updates from identical model and optimizer states, sharing cached teacher targets. Positive-only zeros negative weights while retaining the original denominator and positive weights. Across these eight updates, we measure outcome-aligned changes in observed-token likelihood,
\begin{equation}
    A_i^{(b)}=(2R_i-1)\frac{1}{T_i}\sum_{t=1}^{T_i}
    \log\frac{p_{\theta'_b}(y_{i,t}\mid x_i,y_{i,<t})}
    {p_{\theta}(y_{i,t}\mid x_i,y_{i,<t})},
    \quad b\in\{\mathrm{Full},\mathrm{Positive}\}.
    \label{eq:app-one-step-alignment}
\end{equation}
Positive $A_i$ indicates increased support for correct trajectories or decreased support for incorrect ones. We average sequence means equally, in nats/token. For ordinary trajectories with $|s_i|>10^{-3}$, C1--C4 denote (correct, $s_i>0$), (incorrect, $s_i<0$), (correct, $s_i<0$), and (incorrect, $s_i>0$), respectively. Positive/negative stance means that PI increases/decreases teacher likelihood relative to the no-PI condition. Thus, C1--C2 are outcome-consistent and C3--C4 are outcome-inconsistent.

\begingroup
\begin{figure}[H]
\centering
\includegraphics[width=0.85\linewidth]{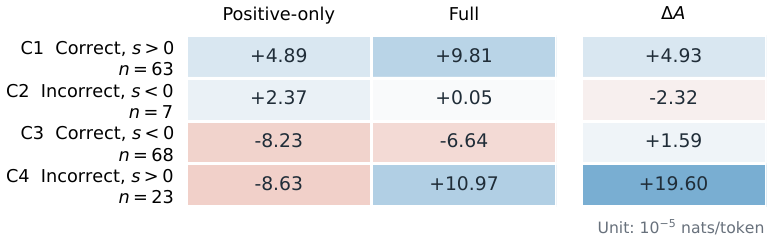}
\caption{\textbf{One-step effects of weak negative supervision.} Mean alignment for ordinary trajectories with $|s_i|>10^{-3}$, grouped by outcome and stance (Eq.~\ref{eq:tau-stance}); $n$ denotes group size. $\Delta A=A^{(\mathrm{Full})}-A^{(\mathrm{Positive})}$. Colors are centered at zero, and positive differences favor Full.}
\label{fig:stance_gate_one_step}
\end{figure}
\endgroup

\paragraph{Evidence for weak negative supervision.}
Across all 256 trajectories, Full improves mean alignment in three of four batches, with a pooled gain of $3.19\times10^{-5}$ nats/token. As shown in Figure~\ref{fig:stance_gate_one_step}, C4 (incorrect, PI-supported) shows the strongest gain, changing from $-8.63$ to $+10.97$ (in $10^{-5}$ nats/token), with positive mean paired differences in all four batches. These combined-update likelihood diagnostics complement the long-horizon accuracy gain from 54.4\% to 55.9\% with weak negative weighting (Figure~\ref{fig:sensitivity_generalization}(d)).

\subsubsection{Truthfulness Substitution across Model Scales}
\label{app:truthfulness-scaling}

\begin{figure}[htbp]
\centering
\includegraphics[width=\linewidth]{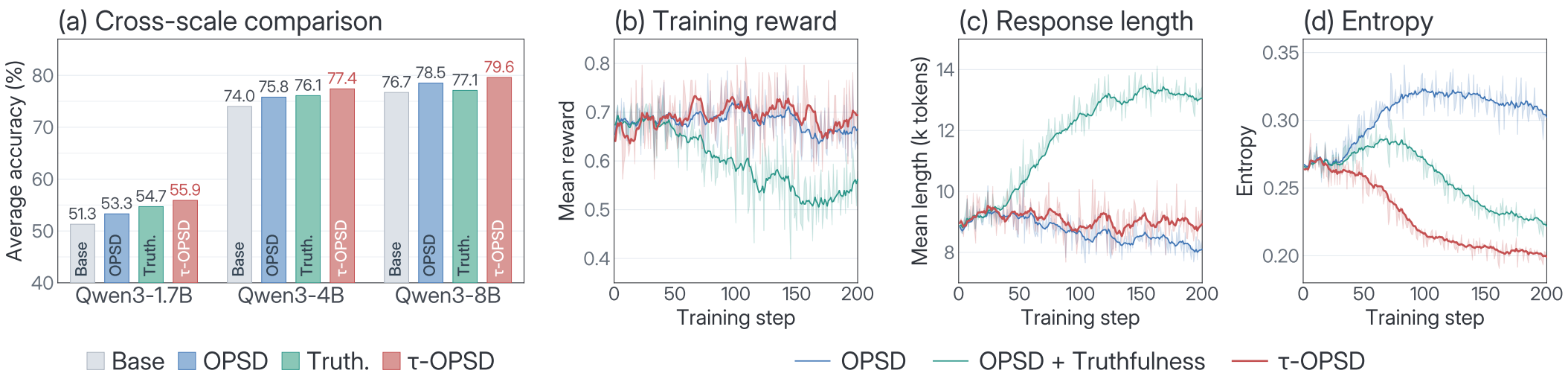}
\caption{\textbf{Truthfulness substitution and scaling.} (a) Mathematical macro-average $\mathrm{avg}@12$ at 1.7B, 4B, and 8B. ``Truth.'' denotes OPSD with solution PI replaced by truthfulness PI. (b--d) Reward, response length, and policy entropy at 1.7B. Thick curves show trailing ten-step means; faint traces and shading show raw observations and their deviations from these means.}
\label{fig:truthfulness_scaling}
\end{figure}

Figure~\ref{fig:truthfulness_scaling} complements the trait comparisons in Section~\ref{sec:traits_beyond_truthfulness} by comparing solution-conditioned OPSD, direct truthfulness substitution within OPSD, and $\tau$-OPSD across model scales. Direct substitution changes only the teacher's PI, whereas $\tau$-OPSD additionally adapts supervision through the proposed framework.

\paragraph{Cross-scale performance.}
At 1.7B, direct truthfulness substitution improves accuracy from 53.3 to 54.7. Its advantage over solution-conditioned OPSD shrinks to 0.3 points at 4B and becomes a 1.4-point deficit at 8B. $\tau$-OPSD achieves the highest accuracy among the compared methods at all three scales, reaching 55.9, 77.4, and 79.6, respectively. These results exceed direct substitution by 1.2, 1.3, and 2.5 percentage points. Thus, behavioral PI provides useful supervision, while the proposed framework consistently outperforms both solution-conditioned OPSD and direct PI substitution across model scales.

\paragraph{Reward and response length.}
At 1.7B, direct substitution shows declining reward and increasing response length after approximately 40 steps. Over the final 20 steps, $\tau$-OPSD achieves mean reward 0.688 with 8.93k-token responses, compared with 0.546 and 13.10k for direct substitution. Solution-conditioned OPSD produces shorter responses but has lower reward than $\tau$-OPSD. Thus, $\tau$-OPSD maintains productive reasoning without the length inflation induced by direct truthfulness substitution.

\paragraph{Entropy dynamics.}
OPSD's entropy rises and remains elevated; direct substitution first increases entropy, then reduces it alongside reward deterioration. $\tau$-OPSD instead exhibits a gradual entropy decline with sustained reward and bounded response length. This joint pattern supports more concentrated predictions while maintaining training performance, linking the benefits of behavioral PI to the supervision mechanism used to distill it.

\subsection{Trait-Specific Privileged Instructions}
\label{app:trait-pi}

We provide the exact privileged instructions used to examine variation in behavioral traits, prompt wording, and semantic content. Appendix~\ref{app:trait-variants} presents planning and exploration instructions for evaluating behavioral PI beyond truthfulness. Appendix~\ref{app:trait_prompts} presents explicit truthfulness and integrity truthfulness instructions for assessing sensitivity to how truthfulness is expressed. Appendix~\ref{app:truthfulness-semantic-prompts} presents the three semantic controls used in the fine-grained truthfulness ablation. All variants guide the teacher on student-generated prefixes without supplying reference answers or external solutions.

\subsubsection{Behavioral Trait Variants}
\label{app:trait-variants}

\paragraph{Downside-aware planning.}
The planning trait encourages early identification of consequential hazards and uncertainty, with caution calibrated to the stakes of the decision. It favors relevant mitigations and reversible intermediate steps before irreversible commitments. The instruction explicitly discourages generic risk lists and indiscriminate caution, so that attention to potential downsides supports continued progress toward a justified solution.

\begin{privilegedinstruction}[colback=orange!3!white,
    colframe=orange!45!black,
    colbacktitle=orange!12!white]{Planning PI: Downside-aware planning}
Please approach the problem with a considered plan. Identify the important constraints, uncertainties, and possible failure points before committing to a path. Organize the work into manageable steps, check assumptions that could invalidate later reasoning, and prefer steps that can be revised before relying on irreversible conclusions. Match the amount of caution to the consequences of being wrong, address the risks that matter for this problem, and keep making progress toward a justified solution rather than producing generic risk lists or becoming stuck in excessive caution.
\end{privilegedinstruction}

\paragraph{Controlled exploration.}
The exploration trait encourages widening the option space when doing so is useful, considering materially distinct plausible paths, and comparing them using task-relevant evidence. It balances this search with an explicit stopping principle: once a path is justified, the teacher should favor convergence rather than redundant branching or continued switching. Thus, the instruction promotes purposeful exploration while discouraging open-ended brainstorming.

\begin{privilegedinstruction}[colback=teal!3!white,
    colframe=teal!45!black,
    colbacktitle=teal!10!white]{Exploration PI: Controlled exploration}
Please explore the problem with an open but disciplined mind. When the path forward is uncertain, consider genuinely different plausible approaches rather than superficial variations, and compare them using the information and reasoning relevant to the problem. Follow the approach best supported by that evidence, and reconsider it when a concrete obstacle or new evidence warrants a change. Once a path is sufficiently justified, carry it through to a conclusion instead of continuing to brainstorm, repeat branches, or switch methods without a reason.
\end{privilegedinstruction}

\subsubsection{Truthfulness Prompt Variants}
\label{app:trait_prompts}

The following variants probe sensitivity to the wording and framing of truthfulness PI (Figure~\ref{fig:sensitivity_generalization}). They retain the emphasis on justified reasoning and uncertainty checking while expressing it through different instructions.

\paragraph{Explicit truthfulness.}
This variant explicitly asks the teacher to evaluate the student's next-token distribution as an honest and epistemically careful reasoner. It specifies preferences for justified claims, checking uncertain steps, avoiding fabrication, and correcting overconfidence, and explicitly prohibits revealing an answer or external solution.

\begin{privilegedinstruction}[colback=violet!3!white,
    colframe=violet!45!black,
    colbacktitle=violet!10!white]{Explicit Truthfulness PI}
Privileged instruction for the teacher:\newline
Evaluate the student's next-token distribution as an honest and epistemically careful reasoner.\newline
Prefer continuations that state only what is justified, check uncertain steps, avoid fabricating facts or unsupported conclusions, and correct overconfident reasoning.\newline
Do not reveal any answer or external solution.
\end{privilegedinstruction}

\noindent\begin{minipage}{\linewidth}
\paragraph{Integrity truthfulness.}
This variant paraphrases the direct problem-solving instruction around integrity and evidence-grounded reasoning. It discourages bluffing, misleading claims, and unproved assumptions, while encouraging verification, acknowledgment of uncertainty, and revision of mistakes. It tests robustness to alternative truthfulness wording without explicitly describing next-token evaluation.
\par\vspace{8pt}
\begin{privilegedinstruction}[before skip=0pt,
    colback=olive!3!white,
    colframe=olive!45!black,
    colbacktitle=olive!12!white]{Integrity Truthfulness PI}
Approach the problem with integrity. Do not bluff, mislead, or take shortcuts that assume what has not been proved. Ground each claim in the given information and sound reasoning, verify any doubtful step, acknowledge uncertainty rather than inventing support, and revise mistakes before presenting the final answer.
\end{privilegedinstruction}
\end{minipage}

\subsubsection{Truthfulness Semantic Controls}
\label{app:truthfulness-semantic-prompts}

These instructions probe the content of truthfulness PI in the fine-grained ablation (Section~\ref{sec:traits_beyond_truthfulness}, Table~\ref{tab:truthfulness_semantic_ablation}). They replace only the teacher-side instruction on ordinary trajectories, retaining local-progress guidance for hard samples and the remaining training configuration.

\paragraph{Generic Reasoning.}
This control emphasizes structured problem solving without explicitly requesting honesty, uncertainty handling, verification, or self-correction.

\begin{privilegedinstruction}{Generic Reasoning PI}
Please solve the problem carefully and systematically. Break the task into manageable steps, identify the relevant quantities and constraints, carry out the necessary derivations or calculations in a clear order, and combine the intermediate results before giving the final answer.
\end{privilegedinstruction}

\paragraph{Verification.}
This control retains calculation and consistency checks without the broader honesty and anti-fabrication framing.

\begin{privilegedinstruction}{Verification PI}
Please solve the problem with explicit verification. After each important derivation or calculation, check that the operation is valid and that the intermediate result is consistent with the problem conditions. Recompute doubtful quantities when needed before giving the final answer.
\end{privilegedinstruction}

\paragraph{Presentation.}
This control emphasizes notation, organization, and concise expression. This control tests whether clearer presentation and reduced repetition provide useful supervision.

\begin{privilegedinstruction}{Presentation PI}
Please present the solution clearly and professionally. Use consistent notation, keep the reasoning well organized, avoid unnecessary repetition, and make the progression between steps easy to follow. State the final answer in a concise and unambiguous form.
\end{privilegedinstruction}

\subsection{Case Study}
\label{app:case-study}

We compare paired \textit{Qwen3-4B} Base and $\tau$-OPSD rollouts on one AMC23 problem and one AIME24 problem, using the trained checkpoint at step 120. Each box follows the opening reasoning, selected checks, and final response, illustrating behaviors encouraged by truthfulness PI (Section~\ref{sec:tau-guidance}). Original wording is retained, with mathematical notation reformatted and emphasis added; italic labels are editorial, and $[\ldots]$ marks omissions.

\subsubsection{AMC23: Revising a Transformation Pattern}
\label{app:case-amc}

\paragraph{Problem.}
Each $T_k$ rotates the plane $k$ degrees counterclockwise about the origin, then reflects it across the $y$-axis. Find the least positive $n$ for which applying $T_1,T_2,\ldots,T_n$ returns $(1,0)$ to itself. The correct answer is $359$.

\begin{privilegedinstruction}[colback=black!2!white,
    colframe=black!40!white,
    colbacktitle=black!7!white]{Base model --- incorrect answer $80$}
\textit{Reasoning excerpt.}
Okay, so I need to figure out the least positive integer $n$ such that performing the sequence of transformations $T_1,T_2,\ldots,T_n$ brings the point $(1,0)$ back to itself. $[\ldots]$

If each $T_k$ is multiplying by $-e^{ik^\circ}$, then applying $T_1,T_2,\ldots,T_n$ would be multiplying by $(-e^{i1^\circ})(-e^{i2^\circ})\cdots(-e^{in^\circ})$.
Which is $(-1)^n e^{i(1^\circ+2^\circ+\cdots+n^\circ)}$. $[\ldots]$

But wait, let me check if $n=80$ satisfies the parity condition.
Earlier, we had $m=n(n+1)/360=80\cdot81/360=(80/360)\cdot81=(2/9)\cdot81=18$. $[\ldots]$ $n=80$, which is even. $m=18$, which is even. Therefore, same parity. $[\ldots]$

\textit{Final response.}
Reflection across the $y$-axis: This corresponds to \textbf{multiplying the complex number by $-1$}. $[\ldots]$ Thus, $n=80$ is the smallest such integer. $[\ldots]$ Final Answer: $\boxed{80}$.
\end{privilegedinstruction}

\begin{privilegedinstruction}[colback=teal!2!white,
    colframe=teal!45!black,
    colbacktitle=teal!9!white]{$\tau$-OPSD --- correct answer $359$}
\textit{Reasoning excerpt.}
Okay, let's see. I need to find the smallest positive integer $n$ such that applying the transformations $T_1,T_2,\ldots,T_n$ brings the point $(1,0)$ back to itself. $[\ldots]$

Wait, matrix multiplication order is important here. Since we first rotate and then reflect, the overall transformation is reflection $*$ rotation. So, if we have a rotation matrix $R$ and reflection matrix $F$, then $T_k$ is $F*R$. $[\ldots]$

Then reflection across the $y$-axis is complex conjugation, right? $[\ldots]$ Wait, no. $[\ldots]$ Suppose $z=x+iy$. Reflect over $y$-axis: $(-x,y)=-x+iy$. $[\ldots]$ So if I want to get $-x+iy$, \textbf{that is equal to $-\overline{z}$}. $[\ldots]$

So the smallest $n$ is $361$. $[\ldots]$

Let me re-examine the earlier calculations.
For $n=1$, applying $T_1$ to $(1,0)$ gives $(-\cos 1^\circ,\sin 1^\circ)$, not $(1,0)$. \textbf{So the pattern I thought was incorrect.} $[\ldots]$

If $n$ is even: $(\cos(k^\circ),-\sin(k^\circ))$, where $k=n/2$.
If $n$ is odd: $(-\cos(k^\circ),\sin(k^\circ))$, where $k=(n+1)/2$. $[\ldots]$

For $n=359$, $k=(359+1)/2=180^\circ$. So the point is
$(-\cos(180^\circ),\sin(180^\circ))=(-(-1),0)=(1,0)$. Correct.
So, the smallest $n$ is $359$, which is smaller than $720$. $[\ldots]$

\textit{Final response.}
Between the two cases, the smallest $n$ is: $\boxed{359}$.
This is the least positive integer such that applying the sequence $T_1,T_2,\ldots,T_n$ returns the point $(1,0)$ to itself.
\end{privilegedinstruction}

\paragraph{Analysis.}
Base uses $z\mapsto-z$ instead of the reflection $z\mapsto-\overline{z}$, so its parity check verifies the wrong transformation. The trained rollout corrects its reflection rule, then rejects its proposed answer $361$ using the case $n=1$. Revising the odd-case index to $(n+1)/2$ yields $359$, compared with $720$ for the even case. Concrete checks change both the governing expressions and the answer.

\newpage
\subsubsection{AIME24: Checking Minimality Beyond a Valid Solution}
\label{app:case-aime}

\paragraph{Problem.}
Let $p$ be the least prime for which $p^2$ divides $n^4+1$ for some positive integer $n$. Find the least positive $m$ such that $p^2$ divides $m^4+1$. Both rollouts identify $p=17$; the correct minimum is $m=110$.

\begin{privilegedinstruction}[colback=black!2!white,
    colframe=black!40!white,
    colbacktitle=black!7!white]{Base model --- incorrect answer $155$}
\textit{Reasoning excerpt.}
Okay, so I need to find the least prime number $p$ such that there's a positive integer $n$ where $n^4+1$ is divisible by $p^2$. Then, once I find that $p$, I have to find the smallest positive integer $m$ such that $m^4+1$ is divisible by $p^2$. Alright, let's start by understanding the problem step by step. $[\ldots]$

Therefore, the first part of the problem is $p=17$. $[\ldots]$ First, we started with $n=2+17k$. $[\ldots]$

However, maybe there is another solution? Wait, Hensel's lemma says that if the derivative is not $0$ mod $p$, then there is a unique lift. Since we had $f'(2)=4\cdot2^3=32\equiv15\pmod{17}\ne0$, so unique lift. \textbf{Therefore, only one solution mod $17^2$.} $[\ldots]$

Therefore, $155$ is the only solution. Therefore, $155$ is the minimal positive integer. $[\ldots]$

\textit{Final response.}
Thus, the smallest $k$ is $9$, and so: $m=2+17\cdot9=155$.
We now verify: $155^4+1\equiv0\pmod{289}$.
We already did the lifting step, and the result is correct. Therefore, the minimal $m$ is: $\boxed{155}$.
\end{privilegedinstruction}

\begin{privilegedinstruction}[colback=teal!2!white,
    colframe=teal!45!black,
    colbacktitle=teal!9!white]{$\tau$-OPSD --- correct answer $110$}
\textit{Reasoning excerpt.}
Okay, so I need to find the least prime number $p$ such that there's a positive integer $n$ where $n^4+1$ is divisible by $p^2$. Then, once I find that $p$, I have to find the smallest positive integer $m$ where $m^4+1$ is divisible by $p^2$. Alright, let's break this down. $[\ldots]$

Wait, but earlier I thought $6^4\equiv16$? \textbf{Wait, that was wrong.} Let me recalculate: Wait, $6^4$ is $6^2$ squared. $6^2=36\equiv2\pmod{17}$. Then $2^2=4$. Therefore, $6^4\equiv4\pmod{17}$. So that's different from my previous calculation. $[\ldots]$

However, in our case, Hensel's lemma gave us one solution. However, since the equation has $4$ solutions, there are $4$ solutions mod $17^2$. Therefore, the minimal positive integer $m$ is the minimal among these $4$ solutions. $[\ldots]$

Alternatively, since we know that $8$ is a solution mod $17$, and we found $110$ mod $17^2$. $[\ldots]$

Therefore, the four solutions are $110$, $179$, $-110=179$, $-179=110$? \textbf{Wait, no.} $[\ldots]$

If $x^4\equiv-1$, then $(x^3)^4=x^{12}=(x^4)^3\equiv(-1)^3=-1\pmod{289}$. Therefore, yes, $x^3$ is also a solution. Therefore, $110^3\bmod289$ is another solution. Therefore, $155$ is another solution. $[\ldots]$ Let me compute $-155\bmod289$. $-155+289=134$. $[\ldots]$

Wait, $110$ is the smallest? $110$ is less than $134$, $155$, $179$. Therefore, $110$ is the minimal. $[\ldots]$

\textit{Final response.}
So, the lifted solution is: $x=8+6\cdot17=8+102=110$.
Thus, the least positive integer $m$ such that $m^4+1\equiv0\pmod{289}$ is: $\boxed{110}$.
\end{privilegedinstruction}

\paragraph{Analysis.}
Base finds a valid root, $155$, but confuses a unique lift of one residue modulo $17$ with a unique solution modulo $289$. Its final verification establishes validity, leaving minimality unresolved. The trained rollout corrects a modular arithmetic error, then recognizes the need for four distinct roots. It notices that negation has reproduced the same pair and recovers the missing pair before comparing $110,134,155,179$. Its checks address both local calculations and the claim required by the question.

Both cases exhibit the testing of uncertain claims and revision of unsupported conclusions encouraged by truthfulness PI. Although Base also uses checking language, the key distinction is whether these checks resolve the decisive error. These examples illustrate how the trained model uses concrete checks to revise intermediate reasoning and reach justified conclusions.

\end{document}

%% file: math_commands.tex
\usepackage{amsmath,amsfonts,bm}

\def\eqref#1{equation~\ref{#1}}

\def\1{\bm{1}}

\DeclareMathAlphabet{\mathsfit}{\encodingdefault}{\sfdefault}{m}{sl}
\SetMathAlphabet{\mathsfit}{bold}{\encodingdefault}{\sfdefault}{bx}{n}













%% file: arxiv.bbl
\begin{thebibliography}{42}
\providecommand{\natexlab}[1]{#1}
\providecommand{\url}[1]{\texttt{#1}}
\expandafter\ifx\csname urlstyle\endcsname\relax
  \providecommand{\doi}[1]{doi: #1}\else
  \providecommand{\doi}{doi: \begingroup \urlstyle{rm}\Url}\fi

\bibitem[Agarwal et~al.(2024)Agarwal, Vieillard, Zhou, Stanczyk, Ramos~Garea, Geist, and Bachem]{agarwal2024gkd}
Rishabh Agarwal, Nino Vieillard, Yongchao Zhou, Piotr Stanczyk, Sabela Ramos~Garea, Matthieu Geist, and Olivier Bachem.
\newblock On-policy distillation of language models: Learning from self-generated mistakes.
\newblock In \emph{The Twelfth International Conference on Learning Representations}, 2024.
\newblock URL \url{https://proceedings.iclr.cc/paper_files/paper/2024/hash/5be69a584901a26c521c2b51e40a4c20-Abstract-Conference.html}.

\bibitem[{AI-MO}(2024{\natexlab{a}})]{aimoAIME}
{AI-MO}.
\newblock {AIMO Validation AIME}.
\newblock Hugging Face dataset, 2024{\natexlab{a}}.
\newblock URL \url{https://huggingface.co/datasets/AI-MO/aimo-validation-aime}.
\newblock Accessed 2026-06-20.

\bibitem[{AI-MO}(2024{\natexlab{b}})]{aimoAMC}
{AI-MO}.
\newblock {AIMO Validation AMC}.
\newblock Hugging Face dataset, 2024{\natexlab{b}}.
\newblock URL \url{https://huggingface.co/datasets/AI-MO/aimo-validation-amc}.
\newblock Accessed 2026-06-20.

\bibitem[Chen et~al.(2025)Chen, Arditi, Sleight, Evans, and Lindsey]{chen2025personavectors}
Runjin Chen, Andy Arditi, Henry Sleight, Owain Evans, and Jack Lindsey.
\newblock Persona vectors: Monitoring and controlling character traits in language models, 2025.
\newblock URL \url{https://arxiv.org/abs/2507.21509}.

\bibitem[Cloud et~al.(2026)Cloud, Le, Chua, Betley, Sztyber-Betley, Mindermann, Hilton, Marks, and Evans]{cloud2026subliminal}
Alex Cloud, Minh Le, James Chua, Jan Betley, Anna Sztyber-Betley, S{\"o}ren Mindermann, Jacob Hilton, Samuel Marks, and Owain Evans.
\newblock Language models transmit behavioural traits through hidden signals in data.
\newblock \emph{Nature}, 652:\penalty0 615--621, 2026.
\newblock \doi{10.1038/s41586-026-10319-8}.
\newblock URL \url{https://www.nature.com/articles/s41586-026-10319-8}.

\bibitem[Cobbe et~al.(2021)Cobbe, Kosaraju, Bavarian, Chen, Jun, Kaiser, Plappert, Tworek, Hilton, Nakano, Hesse, and Schulman]{cobbe2021gsm8k}
Karl Cobbe, Vineet Kosaraju, Mohammad Bavarian, Mark Chen, Heewoo Jun, Lukasz Kaiser, Matthias Plappert, Jerry Tworek, Jacob Hilton, Reiichiro Nakano, Christopher Hesse, and John Schulman.
\newblock Training verifiers to solve math word problems, 2021.
\newblock URL \url{https://arxiv.org/abs/2110.14168}.

\bibitem[Feng et~al.(2026)Feng, Zhao, Zhong, Huang, Gu, Kong, Feng, and Qin]{feng2026persona}
Xiachong Feng, Liang Zhao, Weihong Zhong, Yichong Huang, Yuxuan Gu, Lingpeng Kong, Xiaocheng Feng, and Bing Qin.
\newblock {PERSONA}: Dynamic and compositional inference-time personality control via activation vector algebra.
\newblock In \emph{The Fourteenth International Conference on Learning Representations}, 2026.
\newblock URL \url{https://proceedings.iclr.cc/paper_files/paper/2026/hash/f66ce276436c4cd151535b03841ac713-Abstract-Conference.html}.

\bibitem[Gu et~al.(2024)Gu, Dong, Wei, and Huang]{gu2024minillm}
Yuxian Gu, Li~Dong, Furu Wei, and Minlie Huang.
\newblock {MiniLLM}: Knowledge distillation of large language models.
\newblock In \emph{The Twelfth International Conference on Learning Representations}, 2024.
\newblock URL \url{https://proceedings.iclr.cc/paper_files/paper/2024/hash/8ac015d409635f196f9e3e9dcfb9a94e-Abstract-Conference.html}.

\bibitem[Guo et~al.(2025)Guo, Yang, Zhang, Song, Wang, Zhu, Xu, Zhang, Ma, Bi, Zhang, Yu, Wu, Wu, Gou, Shao, Li, Gao, et~al.]{guo2025deepseekr1}
Daya Guo, Dejian Yang, Haowei Zhang, Junxiao Song, Peiyi Wang, Qihao Zhu, Runxin Xu, Ruoyu Zhang, Shirong Ma, Xiao Bi, Xiaokang Zhang, Xingkai Yu, Yu~Wu, Z.~F. Wu, Zhibin Gou, Zhihong Shao, Zhuoshu Li, Ziyi Gao, et~al.
\newblock {DeepSeek-R1} incentivizes reasoning in {LLMs} through reinforcement learning.
\newblock \emph{Nature}, 645:\penalty0 633--638, 2025.
\newblock \doi{10.1038/s41586-025-09422-z}.
\newblock URL \url{https://www.nature.com/articles/s41586-025-09422-z}.

\bibitem[He et~al.(2026)He, Liang, Xu, Liu, Chen, Wang, Song, Yu, Liang, Wang, Zhang, Wang, Tu, Mi, and Yu]{he2025deepmath}
Zhiwei He, Tian Liang, Jiahao Xu, Qiuzhi Liu, Xingyu Chen, Yue Wang, Linfeng Song, Dian Yu, Zhenwen Liang, Wenxuan Wang, Zhuosheng Zhang, Rui Wang, Zhaopeng Tu, Haitao Mi, and Dong Yu.
\newblock Deepmath-103k: A large-scale, challenging, decontaminated, and verifiable mathematical dataset for advancing reasoning.
\newblock In \emph{The Fourteenth International Conference on Learning Representations}, 2026.
\newblock URL \url{https://openreview.net/forum?id=kHB5Te5IWm}.

\bibitem[Hendrycks et~al.(2021)Hendrycks, Burns, Basart, Zou, Mazeika, Song, and Steinhardt]{hendrycks2021mmlu}
Dan Hendrycks, Collin Burns, Steven Basart, Andy Zou, Mantas Mazeika, Dawn Song, and Jacob Steinhardt.
\newblock Measuring massive multitask language understanding.
\newblock In \emph{International Conference on Learning Representations}, 2021.
\newblock URL \url{https://openreview.net/forum?id=d7KBjmI3GmQ}.

\bibitem[Huang et~al.(2026)Huang, Chen, Luo, Dai, Hu, and Zhao]{huang2026sgsd}
Jiazhen Huang, Xiao Chen, Xiao Luo, Yong Dai, Senkang Hu, and Yuzhi Zhao.
\newblock Skill-conditioned gated self-distillation for llm reasoning, 2026.
\newblock URL \url{https://arxiv.org/abs/2605.28791}.

\bibitem[Huang et~al.(2024)Huang, Chen, Mishra, Zheng, Yu, Song, and Zhou]{huang2024selfcorrect}
Jie Huang, Xinyun Chen, Swaroop Mishra, Huaixiu~Steven Zheng, Adams Yu, Xinying Song, and Denny Zhou.
\newblock Large language models cannot self-correct reasoning yet.
\newblock In \emph{The Twelfth International Conference on Learning Representations}, 2024.
\newblock URL \url{https://proceedings.iclr.cc/paper_files/paper/2024/hash/8b4add8b0aa8749d80a34ca5d941c355-Abstract-Conference.html}.

\bibitem[H{\"u}botter et~al.(2026)H{\"u}botter, L{\"u}beck, Behric, Baumann, Bagatella, Marta, Hakimi, Shenfeld, Kleine~Buening, Guestrin, and Krause]{hubotter2026sdpo}
Jonas H{\"u}botter, Frederike L{\"u}beck, Lejs Behric, Anton Baumann, Marco Bagatella, Daniel Marta, Ido Hakimi, Idan Shenfeld, Thomas Kleine~Buening, Carlos Guestrin, and Andreas Krause.
\newblock Reinforcement learning via self-distillation.
\newblock In \emph{Proceedings of the 43rd International Conference on Machine Learning}, 2026.
\newblock URL \url{https://openreview.net/forum?id=QkfkxyRizZ}.

\bibitem[Jagadeesh et~al.(2026)Jagadeesh, Arora, Saab, Malik, Trofimov, Tsimpourlas, Heidecke, and Singhal]{jagadeesh2026beneficial}
Akshay~V. Jagadeesh, Rahul~K. Arora, Khaled Saab, Ali Malik, Mikhail Trofimov, Foivos Tsimpourlas, Johannes Heidecke, and Karan Singhal.
\newblock Reinforcement learning towards broadly and persistently beneficial models.
\newblock arXiv preprint arXiv:2606.24014, 2026.
\newblock URL \url{https://arxiv.org/abs/2606.24014}.

\bibitem[Jin et~al.(2026)Jin, Min, Yang, Wei, Zhou, Kadhe, Baracaldo, and Lee]{jin2026eopd}
Woogyeol Jin, Taywon Min, Yongjin Yang, Dennis Wei, Yi~Zhou, Swanand~Ravindra Kadhe, Nathalie Baracaldo, and Kimin Lee.
\newblock Entropy-aware on-policy distillation of language models.
\newblock In \emph{Proceedings of the 43rd International Conference on Machine Learning}, 2026.
\newblock URL \url{https://research.ibm.com/publications/entropy-aware-on-policy-distillation-of-language-models--1}.

\bibitem[Kirichenko et~al.(2025)Kirichenko, Ibrahim, Chaudhuri, and Bell]{kirichenko2025abstentionbench}
Polina Kirichenko, Mark Ibrahim, Kamalika Chaudhuri, and Samuel~J. Bell.
\newblock {AbstentionBench}: Reasoning {LLMs} fail on unanswerable questions.
\newblock In \emph{Advances in Neural Information Processing Systems}, volume~38, 2025.
\newblock \doi{10.52202/085713-5729}.
\newblock URL \url{https://papers.nips.cc/paper_files/paper/2025/hash/fb122bfc3f0127a94ded048b5b03496f-Abstract-Datasets_and_Benchmarks_Track.html}.
\newblock Datasets and Benchmarks Track.

\bibitem[Kumar et~al.(2025)Kumar, Zhuang, Agarwal, Su, Co-Reyes, Singh, Baumli, Iqbal, Bishop, Roelofs, Zhang, McKinney, Shrivastava, Paduraru, Tucker, Precup, Behbahani, and Faust]{kumar2025score}
Aviral Kumar, Vincent Zhuang, Rishabh Agarwal, Yi~Su, John~D. Co-Reyes, Avi Singh, Kate Baumli, Shariq Iqbal, Colton Bishop, Rebecca Roelofs, Lei Zhang, Kay McKinney, Disha Shrivastava, Cosmin Paduraru, George Tucker, Doina Precup, Feryal Behbahani, and Aleksandra Faust.
\newblock Training language models to self-correct via reinforcement learning.
\newblock In \emph{The Thirteenth International Conference on Learning Representations}, 2025.
\newblock URL \url{https://proceedings.iclr.cc/paper_files/paper/2025/hash/871ac99fdc5282d0301934d23945ebaa-Abstract-Conference.html}.

\bibitem[Li et~al.(2026{\natexlab{a}})Li, Yang, Fang, Song, Zheng, Guo, Zhang, Wang, and Chua]{li2026srpo}
Gengsheng Li, Tianyu Yang, Junfeng Fang, Mingyang Song, Mao Zheng, Haiyun Guo, Dan Zhang, Jinqiao Wang, and Tat-Seng Chua.
\newblock Unifying group-relative and self-distillation policy optimization via sample routing.
\newblock arXiv preprint arXiv:2604.02288, 2026{\natexlab{a}}.
\newblock URL \url{https://arxiv.org/abs/2604.02288}.

\bibitem[Li et~al.(2026{\natexlab{b}})Li, Zuo, He, Zhang, Xiao, Qian, Yu, Gao, Yang, Liu, and Ding]{li2026rethinking}
Yaxuan Li, Yuxin Zuo, Bingxiang He, Jinqian Zhang, Chaojun Xiao, Cheng Qian, Tianyu Yu, Huan-ang Gao, Wenkai Yang, Zhiyuan Liu, and Ning Ding.
\newblock Rethinking on-policy distillation of large language models: Phenomenology, mechanism, and recipe.
\newblock arXiv preprint arXiv:2604.13016, 2026{\natexlab{b}}.
\newblock URL \url{https://arxiv.org/abs/2604.13016}.

\bibitem[Lightman et~al.(2024)Lightman, Kosaraju, Burda, Edwards, Baker, Lee, Leike, Schulman, Sutskever, and Cobbe]{lightman2024verify}
Hunter Lightman, Vineet Kosaraju, Yuri Burda, Harrison Edwards, Bowen Baker, Teddy Lee, Jan Leike, John Schulman, Ilya Sutskever, and Karl Cobbe.
\newblock Let's verify step by step.
\newblock In \emph{The Twelfth International Conference on Learning Representations}, 2024.
\newblock URL \url{https://proceedings.iclr.cc/paper_files/paper/2024/hash/aca97732e30bcf1303bc22ac3924fd16-Abstract-Conference.html}.

\bibitem[Lu \& Lab(2025)Lu and Lab]{lu2025onpolicydistillation}
Kevin Lu and Thinking~Machines Lab.
\newblock On-policy distillation.
\newblock \emph{Thinking Machines Lab: Connectionism}, 2025.
\newblock \doi{10.64434/tml.20251026}.
\newblock https://thinkingmachines.ai/blog/on-policy-distillation.

\bibitem[Marks et~al.(2026)Marks, Lindsey, and Olah]{marks2026psm}
Sam Marks, Jack Lindsey, and Christopher Olah.
\newblock The {Persona Selection Model}: Why {AI} assistants might behave like humans.
\newblock Anthropic Alignment Science Blog, February 2026.
\newblock URL \url{https://alignment.anthropic.com/2026/psm/}.

\bibitem[{OpenCompass}(2025)]{opencompassAIME2025}
{OpenCompass}.
\newblock {AIME2025}.
\newblock Hugging Face dataset, 2025.
\newblock URL \url{https://huggingface.co/datasets/opencompass/AIME2025}.
\newblock Dataset release; accessed 2026-09-20.

\bibitem[Pan et~al.(2026)Pan, Tao, Zhai, Zhang, Liu, Ding, Liu, and Wen]{pan2026rlcsd}
Leyi Pan, Shuchang Tao, Yunpeng Zhai, Lingzhe Zhang, Zhaoyang Liu, Bolin Ding, Aiwei Liu, and Lijie Wen.
\newblock {RLCSD}: Reinforcement learning with contrastive on-policy self-distillation.
\newblock arXiv preprint arXiv:2606.11709, 2026.
\newblock URL \url{https://arxiv.org/abs/2606.11709}.

\bibitem[Rein et~al.(2024)Rein, Hou, Stickland, Petty, Pang, Dirani, Michael, and Bowman]{rein2023gpqa}
David Rein, Betty~Li Hou, Asa~Cooper Stickland, Jackson Petty, Richard~Yuanzhe Pang, Julien Dirani, Julian Michael, and Samuel~R. Bowman.
\newblock {GPQA}: A graduate-level google-proof q\&a benchmark.
\newblock In \emph{First Conference on Language Modeling}, 2024.
\newblock URL \url{https://openreview.net/forum?id=Ti67584b98}.

\bibitem[Rezaei et~al.(2026)Rezaei, Mahmoud, Wang, Tyagi, Gosai, Dumitru, Sabharwal, Liu, and He]{rezaei2026rubric}
MohammadHossein Rezaei, Anas Mahmoud, Zihao Wang, Utkarsh Tyagi, Advait Gosai, Razvan-Gabriel Dumitru, Aakash Sabharwal, Bing Liu, and Yunzhong He.
\newblock Rubric-guided self-distillation: Post-training without rubric verifiers.
\newblock arXiv preprint arXiv:2606.12507, 2026.
\newblock URL \url{https://arxiv.org/abs/2606.12507}.

\bibitem[Sang et~al.(2026)Sang, Xu, Zhou, He, Wang, and Sun]{sang2026crisp}
Hejian Sang, Yuanda Xu, Zhengze Zhou, Ran He, Zhipeng Wang, and Jiachen Sun.
\newblock {CRISP}: Compressed reasoning via iterative self-policy distillation.
\newblock arXiv preprint arXiv:2603.05433, 2026.
\newblock URL \url{https://arxiv.org/abs/2603.05433}.

\bibitem[Shao et~al.(2024)Shao, Wang, Zhu, Xu, Song, Bi, Zhang, Zhang, Li, Wu, and Guo]{shao2024deepseekmath}
Zhihong Shao, Peiyi Wang, Qihao Zhu, Runxin Xu, Junxiao Song, Xiao Bi, Haowei Zhang, Mingchuan Zhang, Y.~K. Li, Y.~Wu, and Daya Guo.
\newblock {DeepSeekMath}: Pushing the limits of mathematical reasoning in open language models.
\newblock arXiv preprint arXiv:2402.03300, 2024.
\newblock URL \url{https://arxiv.org/abs/2402.03300}.

\bibitem[Snell et~al.(2025)Snell, Lee, Xu, and Kumar]{snell2025scaling}
Charlie Snell, Jaehoon Lee, Kelvin Xu, and Aviral Kumar.
\newblock Scaling {LLM} test-time compute optimally can be more effective than scaling parameters for reasoning.
\newblock In \emph{The Thirteenth International Conference on Learning Representations}, 2025.
\newblock URL \url{https://proceedings.iclr.cc/paper_files/paper/2025/hash/1b623663fd9b874366f3ce019fdfdd44-Abstract-Conference.html}.

\bibitem[Song et~al.(2026)Song, Chen, Tajwar, Munos, Pathak, Bagnell, Singh, and Zanette]{song2026rltf}
Yuda Song, Lili Chen, Fahim Tajwar, R{\'e}mi Munos, Deepak Pathak, J.~Andrew Bagnell, Aarti Singh, and Andrea Zanette.
\newblock Expanding the capabilities of reinforcement learning via text feedback.
\newblock In \emph{Proceedings of the 43rd International Conference on Machine Learning}, 2026.
\newblock URL \url{https://rl-textfeedback.github.io/}.

\bibitem[Stein et~al.(2026)Stein, Huang, and Goldstein]{stein2026gates}
Alex Stein, Furong Huang, and Tom Goldstein.
\newblock {GATES}: Self-distillation under privileged context with consensus gating, 2026.
\newblock URL \url{https://arxiv.org/abs/2602.20574}.

\bibitem[Tan et~al.(2026)Tan, Woodruff, Warncke, Jose, Rich{\'e}, Africa, and Taylor]{tan2026inoculation}
Daniel Tan, Anders Woodruff, Niels Warncke, Arun Jose, Maxime Rich{\'e}, David~Demitri Africa, and Mia Taylor.
\newblock Inoculation prompting: Eliciting traits from {LLM}s during training can reduce trait expression at test-time.
\newblock In \emph{The Fourteenth International Conference on Learning Representations}, 2026.
\newblock URL \url{https://proceedings.iclr.cc/paper_files/paper/2026/hash/2cddea6e86103826c6ae9c59ece64ff7-Abstract-Conference.html}.

\bibitem[Wang et~al.(2026)Wang, Wang, Xiao, Zhou, Pan, Wang, Xu, Wen, Ruan, Chen, and Qi]{wang2026skillsd}
Hao Wang, Guozhi Wang, Han Xiao, Yufeng Zhou, Yue Pan, Jichao Wang, Ke~Xu, Yafei Wen, Xiaohu Ruan, Xiaoxin Chen, and Honggang Qi.
\newblock {Skill-SD}: Skill-conditioned self-distillation for multi-turn {LLM} agents, 2026.
\newblock URL \url{https://arxiv.org/abs/2604.10674}.

\bibitem[Yang et~al.(2025)Yang, Li, Yang, Zhang, Hui, Zheng, Yu, Gao, Huang, Lv, et~al.]{yang2025qwen3}
An~Yang, Anfeng Li, Baosong Yang, Beichen Zhang, Binyuan Hui, Bo~Zheng, Bowen Yu, Chang Gao, Chengen Huang, Chenxu Lv, et~al.
\newblock {Qwen3} technical report.
\newblock arXiv preprint arXiv:2505.09388, 2025.
\newblock URL \url{https://arxiv.org/abs/2505.09388}.

\bibitem[Yang et~al.(2026)Yang, Qin, Si, Chen, Gu, Yao, Lin, Wang, Wang, and Duan]{yang2026rlsd}
Chenxu Yang, Chuanyu Qin, Qingyi Si, Minghui Chen, Naibin Gu, Dingyu Yao, Zheng Lin, Weiping Wang, Jiaqi Wang, and Nan Duan.
\newblock Self-distilled {RLVR}, 2026.
\newblock URL \url{https://arxiv.org/abs/2604.03128}.

\bibitem[Ye et~al.(2026{\natexlab{a}})Ye, Wang, Ling, Xiao, Li, Ma, and Hui]{ye2026subnetworks}
Ruimeng Ye, Zihan Wang, Zinan Ling, Yang Xiao, Manling Li, Xiaolong Ma, and Bo~Hui.
\newblock Your language model secretly contains personality subnetworks.
\newblock In \emph{The Fourteenth International Conference on Learning Representations}, 2026{\natexlab{a}}.
\newblock URL \url{https://proceedings.iclr.cc/paper_files/paper/2026/hash/9b2086bc4a12c3a8e80105635fd95f66-Abstract-Conference.html}.

\bibitem[Ye et~al.(2026{\natexlab{b}})Ye, Dong, Wu, Huang, and Wei]{ye2026opcd}
Tianzhu Ye, Li~Dong, Xun Wu, Shaohan Huang, and Furu Wei.
\newblock On-policy context distillation for language models.
\newblock arXiv preprint arXiv:2602.12275, 2026{\natexlab{b}}.
\newblock URL \url{https://arxiv.org/abs/2602.12275}.

\bibitem[Yeo et~al.(2026)Yeo, Choi, Ki, and Hwang]{yeo2026hintsd}
Woongyeong Yeo, Yumin Choi, Taekyung Ki, and Sung~Ju Hwang.
\newblock Hint-sd: Targeted hindsight self-distillation for long-horizon agents, 2026.
\newblock URL \url{https://arxiv.org/abs/2605.17873}.

\bibitem[Yu et~al.(2026)Yu, Li, Zhao, Liu, Zhang, Wang, Luo, Wu, Mittal, Fredrikson, and Hu]{yu2026multirollout}
Weichen Yu, Xiaomin Li, Yizhou Zhao, Xiaoze Liu, Ruowang Zhang, Haixin Wang, Yinyi Luo, Chen~Henry Wu, Gaurav Mittal, Matt Fredrikson, and Yu~Hu.
\newblock Multi-rollout on-policy distillation via peer successes and failures.
\newblock arXiv preprint arXiv:2605.12652, 2026.
\newblock URL \url{https://arxiv.org/abs/2605.12652}.

\bibitem[Zhao et~al.(2026{\natexlab{a}})Zhao, Xie, Liu, Huang, Pang, Chen, and Grover]{zhao2026opsd}
Siyan Zhao, Zhihui Xie, Mengchen Liu, Jing Huang, Guan Pang, Feiyu Chen, and Aditya Grover.
\newblock Self-distilled reasoner: On-policy self-distillation for large language models.
\newblock In \emph{Proceedings of the 43rd International Conference on Machine Learning}, 2026{\natexlab{a}}.
\newblock URL \url{https://icml.cc/virtual/2026/poster/64784}.

\bibitem[Zhao et~al.(2026{\natexlab{b}})Zhao, Ma, Yang, Feng, Shi, He, Xin, Ren, and Wu]{zhao2026rosd}
Ziqi Zhao, Xinyu Ma, Liu Yang, Yujie Feng, Daiting Shi, Jingzhou He, Xin Xin, Zhaochun Ren, and Xiao-Ming Wu.
\newblock {ROSD}: Reflective on-policy self-distillation for language model reasoning across domains, 2026{\natexlab{b}}.
\newblock URL \url{https://arxiv.org/abs/2605.28014}.

\end{thebibliography}
